\documentclass[11pt, a4paper, copyright, gdm]{google}

\usepackage[authoryear, sort&compress, round]{natbib}
\uselogo{} 

\title{LiteFrame: Efficient Vision Encoders Unlock Frame Scaling in Video LLMs}

\author[1,2]{Jihwan Kim}
\author[1]{Nikhil Parthasarathy}
\author[1]{Danfeng Qin}
\author[1]{Junhwa Hur}
\author[1]{Deqing Sun}
\author[1,2]{Bohyung Han}
\author[1]{Ming-Hsuan~Yang}
\author[1]{Boqing Gong}

\affil[1]{\thepa{}{}}
\affil[2]{Seoul National University}

\begin{abstract}
The fundamental challenge in scaling Video Large Language Models~(Video LLMs) to long-form video lies in managing the explosion of visual-token context length.
Existing strategies predominantly focus on ``post-hoc'' token reduction---reducing visual tokens after feature extraction to alleviate the LLM's computational overhead.
While these methods effectively reduce the number of visual tokens, we observe that the primary latency bottleneck then shifts from the LLM to the expensive per-frame processing of the vision encoder.
To address this, we introduce \emph{LiteFrame}, a strong, yet highly efficient video encoder backbone for Video LLMs.
To train LiteFrame, we propose Compressed Token Distillation~(CTD), a novel training framework that teaches a compact student vision encoder to directly predict information-dense, spatio-temporally compressed representations produced by a large teacher vision model, effectively bypassing redundant computation.
When coupled with further Language Model Adaptation~(LMA), this approach results in a new latency-accuracy Pareto frontier---compared with InternVL3-8B, LiteFrame provides a 35\% reduction in end-to-end latency while processing 8$\times$ more frames \emph{and} improves average video understanding accuracy across multiple benchmarks. Our results demonstrate a new potential path to unlocking longer-form video understanding under fixed compute budgets.

\vspace{4mm}
Project Page: \href{https://jjihwan.github.io/projects/LiteFrame}{jjihwan.github.io/projects/LiteFrame}

\end{abstract}
\begin{document}

\maketitle


\section{Introduction}
\label{sec:intro}

\begin{figure}[ht]
    \centering
    \vspace{-3mm}
    \includegraphics[width=\linewidth]{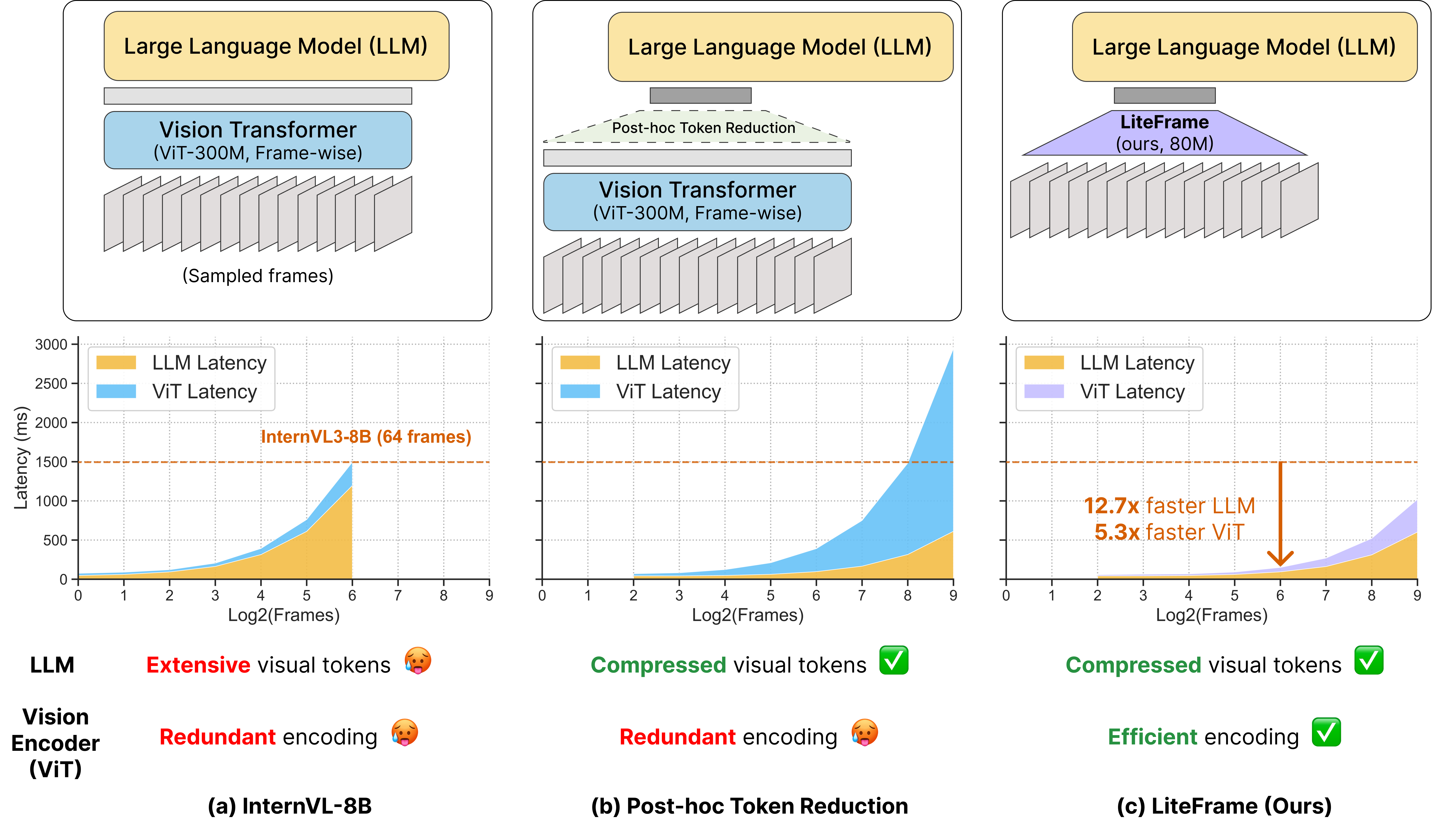}
    \vspace{-8mm}
    \caption{
    We propose \textbf{LiteFrame}, a lightweight video encoder that reduces inefficiencies from both the LLM and the ViT.
    (a) Standard Video LLMs~\citep{zhu2025internvl3} are bottlenecked by the LLM’s quadratic complexity, strictly limiting context length to $\sim$64 frames. (b) Post-hoc reduction alleviates LLM's burden, enabling more frames, yet ironically shifts the bottleneck to the ViT, causing latency to explode. (c) Our approach~(LiteFrame) resolves both inefficiencies, enabling 12.7$\times$ faster LLM prefilling and 5.3$\times$ faster ViT encoding at 64 frames compared to the InternVL3-8B baseline.
    }
    \vspace{-2mm}
\label{fig:stacked_plots_latency}
\end{figure}

Modern Multimodal LLMs~(MLLMs)~\citep{zhu2025internvl3,bai2025qwen2,wang2025internvideo2,li2024llava, pichai2025new} have achieved remarkable progress in recent years in video understanding, parsing complex temporal dynamics for captioning~\citep{fang2024mmbenchvideo}, question answering~\citep{fu2025videomme, zhou2025mlvu}, and reasoning~\citep{fu2026videommev2}.
Despite these strong capabilities, there remains a fundamental scaling problem when handling long-form video within the current paradigm---the computational cost of processing spatio-temporal video data grows prohibitively with increasing frame counts.
To understand why this is, we first note that these models all typically follow a very similar multi-stage architecture consisting of an image encoder (e.g. Vision Transformer; \citealp{dosovitskiy2021vit}) that processes a video frame by frame, an alignment projector, and an LLM that reasons over the interleaved visual and text tokens.
Therefore, with each additional input frame, the computational cost increases due to processing demands in \textit{both} the vision encoder and the LLM.

Existing works that try to alleviate this computational burden have largely focused on the LLM, attributing the primary bottleneck to the quadratic complexity of self-attention over an increasing number of visual tokens.
Consequently, the dominant solution has been an ``extract-and-reduce” paradigm,
which maintains the frozen image encoder for frame-level feature extraction, and leverages post-hoc token reduction strategies (\Cref{fig:stacked_plots_latency} (b))---either spatially~\citep{shang2025llavaprumerge, yang2025visionzip, wang2025dymu}, spatio-temporally~\citep{tao2025dycoke, huang2025prunevid, shen2025fastvid, shao2025holitom}, or via query-guided pruning~\citep{chen2024fastv, xing2025pyramiddrop, shen2024longvu, yang2025topv}---before feeding them to the LLM.

We show that this class of methods ignores the cost of the per-frame feature extraction, which while seemingly lightweight, becomes cumulatively expensive.
Specifically, our preliminary analysis in~\Cref{sec:prelim} reveals that while aggressive post-hoc token reduction~(e.g., $16\times$) alleviates the LLM overhead, as the LLM compute decreases, the computational burden of the visual encoder begins to dominate.
This remaining bottleneck prohibitively sets a floor for the achievable end-to-end inference efficiency.
As illustrated in \Cref{fig:stacked_plots_latency}~(b), once post-hoc token reduction is effectively applied, the vision encoder’s latency becomes the new bottleneck as frame counts increase.
Hence, unlocking the next generation of efficient MLLMs for long-video understanding requires a holistic approach that simultaneously optimizes both visual encoding and language model efficiency.

To this end, we introduce LiteFrame, a lightweight, efficient video encoder designed to reduce per-frame compute with a minimal decrease in video understanding accuracy.
To achieve this, we propose Compressed Token Distillation~(CTD), a novel strategy for training a compute-efficient, token-compressive encoder from a pretrained teacher image encoder.
Specifically, CTD directly aligns the student with an information-dense, spatio-temporally compressed teacher output.
Furthermore, we design the student encoder architecture to explicitly reduce spatio-temporal redundancies across frames. 

\begin{figure}[!t]
    \centering
    \includegraphics[width=0.65\linewidth]{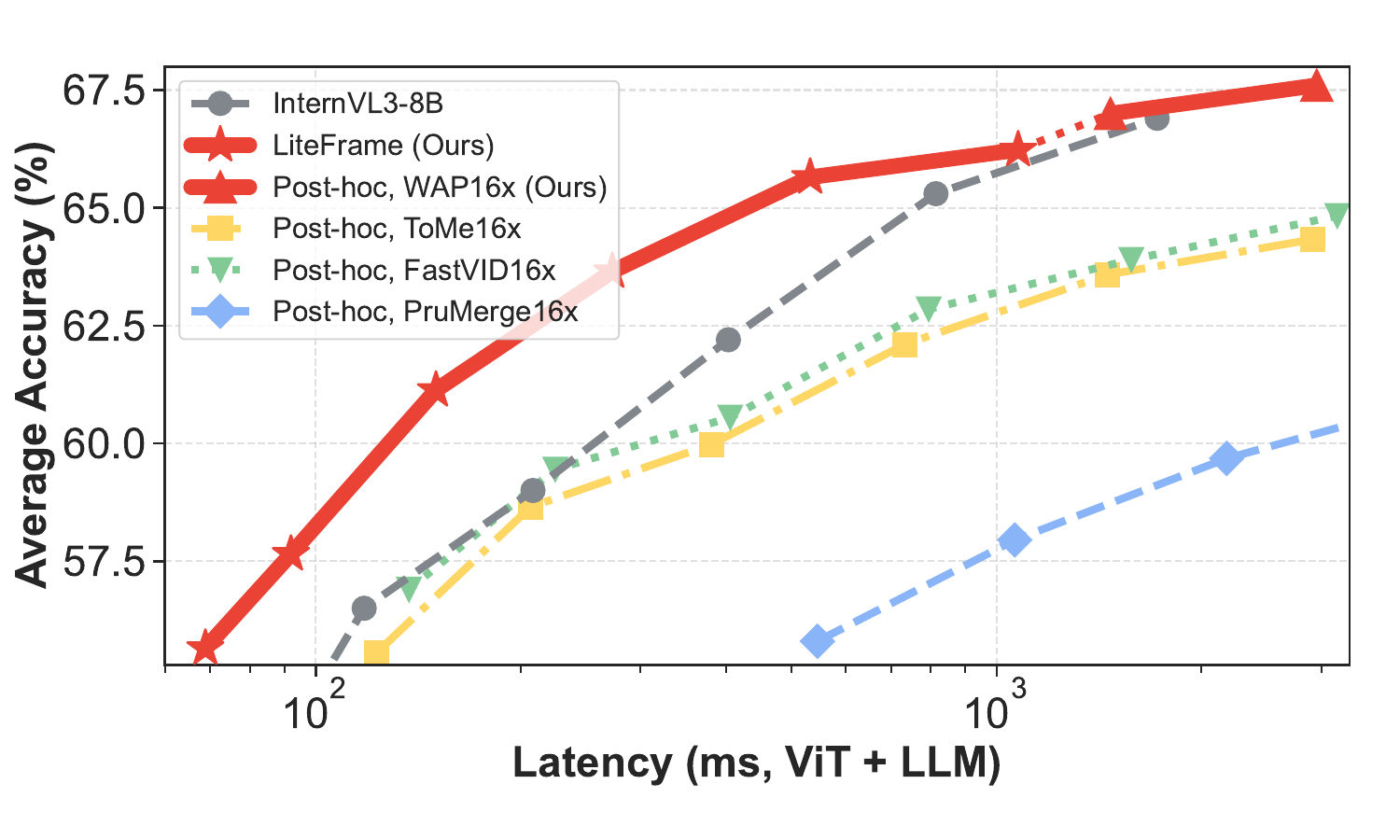}
    \caption{\textbf{The pareto frontier of video understanding efficiency.}
    We illustrate the trade-off between average accuracy on four video benchmarks~(Video-MME w/ and w/o subtitles, MLVU, and LongVideoBench) and end-to-end latency including vision encoding and LLM prefilling.
    Our proposed post-hoc primitive, Weighted Average Pooling~(WAP, red triangles), and our efficient video encoder, LiteFrame~(red stars), push the efficiency Pareto frontier, achieving superior accuracy compared to the teacher model (InternVL3, black dashed).
    Existing post-hoc methods~(color dashed) fail to improve the trade-off as they neglect the encoder latency bottleneck.
    Note that the x-axis~(latency) is log-scaled.
}
\label{fig:teaser}
\end{figure}

When coupled with a lightweight Language Model Adaptation~(LMA) stage~(adapting the new encoder with the LLM), LiteFrame allows Video LLMs to achieve a new latency-accuracy Pareto frontier for video understanding.
As illustrated in~\Cref{fig:teaser}, our model delivers superior accuracy with remarkably low latency when compared to existing baselines.
Specifically, LiteFrame significantly outperforms the InternVL3-8B by processing $8\times$ more frames with a 35\% reduction in end-to-end latency, while using only 87M parameters (vs.\ 304M for the teacher).

To summarize our contributions:
\begin{itemize}
    \item We identify a critical scaling blindspot in current efficient Video LLM paradigms: while post-hoc token reduction effectively alleviates LLM computational costs, the vision encoder becomes the new latency bottleneck, preventing further efficient scaling to long videos.
    
    \item We propose LiteFrame, an efficient video encoder that resolves this bottleneck shift by integrating token compression directly within a lightweight visual backbone.
    
    \item We introduce Compressed Token Distillation (CTD), a novel training framework for maximizing the transfer of spatio-temporally dense information from a teacher to a compact student.
    
    \item Extensive experiments demonstrate that LiteFrame redefines the performance-latency trade-off. Our approach achieves a $1.53\times$ acceleration in end-to-end inference compared to the InternVL3-8B teacher, while processing $8\times$ more frames and outperforming the baselines on multiple video understanding tasks.
\end{itemize}

\section{Related work}
\label{sec:relwork}

\noindent \textbf{Post-hoc token reduction.}
The predominant method for making MLLMs efficient is the ``extract-and-reduce'' paradigm: applying post-hoc token reduction after heavy, pre-trained vision encoders extract dense features, aiming to reduce the cost attributed to the LLM's quadratic complexity.
Early approaches focused on spatial redundancy within individual images, using adaptive selection or merging~\citep{shang2025llavaprumerge, yang2025visionzip}.
More recent efforts extend this to the temporal dimension for video inputs via dynamic pruning or holistic merging~\citep{shen2025fastvid, tao2025dycoke, huang2025prunevid, shao2025holitom}.

While these post-hoc methods reduce the computational burden on the LLM, they remain inefficient for long-form video understanding (hundreds or thousands of frames) because they miss a critical scaling bottleneck.
Because these methods rely on a heavy, frozen encoder to process every frame prior to compression, the latency bottleneck shifts from the LLM to vision encoding.

\noindent \textbf{Efficient vision encoders for MLLMs.}
A parallel line of work aims to reduce the cost of visual encoding.
MobileNet-v5~\citep{google2025gemma3n, qin2024mobilenetv4} achieves high inference throughput on edge devices through aggressive architectural optimization.
FastVLM~\citep{vasu2025fastvlm} introduces FastViTHD, a hybrid encoder that combines convolutional efficiency with transformer-based global modeling to better balance latency and input resolution. However, these methods focus on image-centric architectures that are highly effective for spatial encoding but do not explicitly exploit the strong temporal redundancy across frames.

In the video domain, Video-Panda~\citep{yi2025videopanda} proposes an encoder-free paradigm, using a Spatio-Temporal Alignment Block to bypass a heavy visual backbone. This removes the visual backbone bottleneck but exposes the downstream LLM to dense, uncompressed token streams, shifting the bottleneck back to the LLM.
More recently, AutoGaze~\citep{shi2026autogaze} trains a lightweight module to pre-filter visual tokens before they are processed by the ViT.
While it successfully reduces tokens, this method introduces additional latency overhead, including the cost of a heavy VideoViT and autoregressive decoding within the reduction module, ultimately degrading the latency-accuracy trade-off when evaluated on long videos.
\begin{figure}[!t]
    \centering
    \begin{minipage}[b]{0.48\textwidth}
        \centering
        \includegraphics[width=\linewidth]{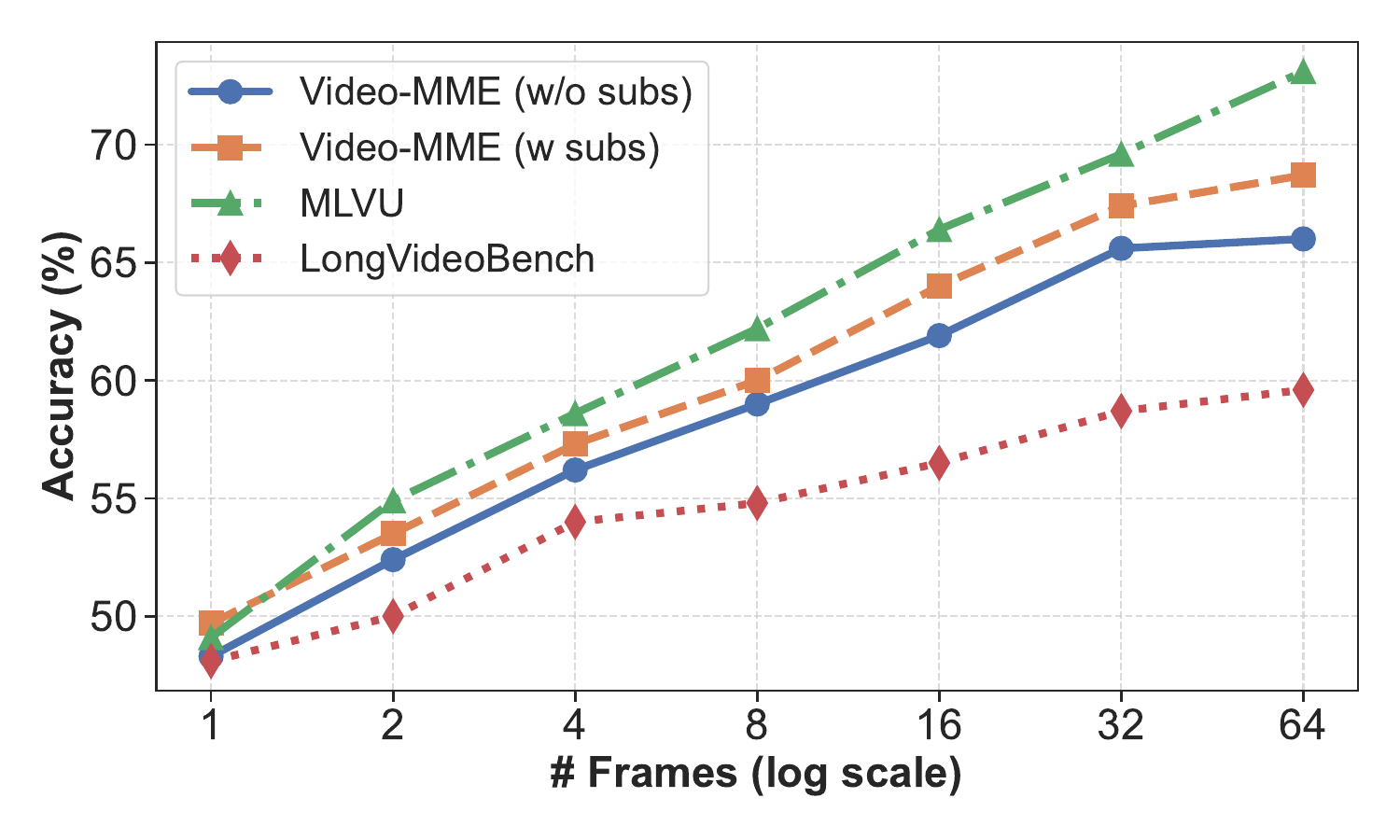} 
    \end{minipage}\hfill
    \begin{minipage}[b]{0.48\textwidth}
        \centering
        \includegraphics[width=\linewidth]{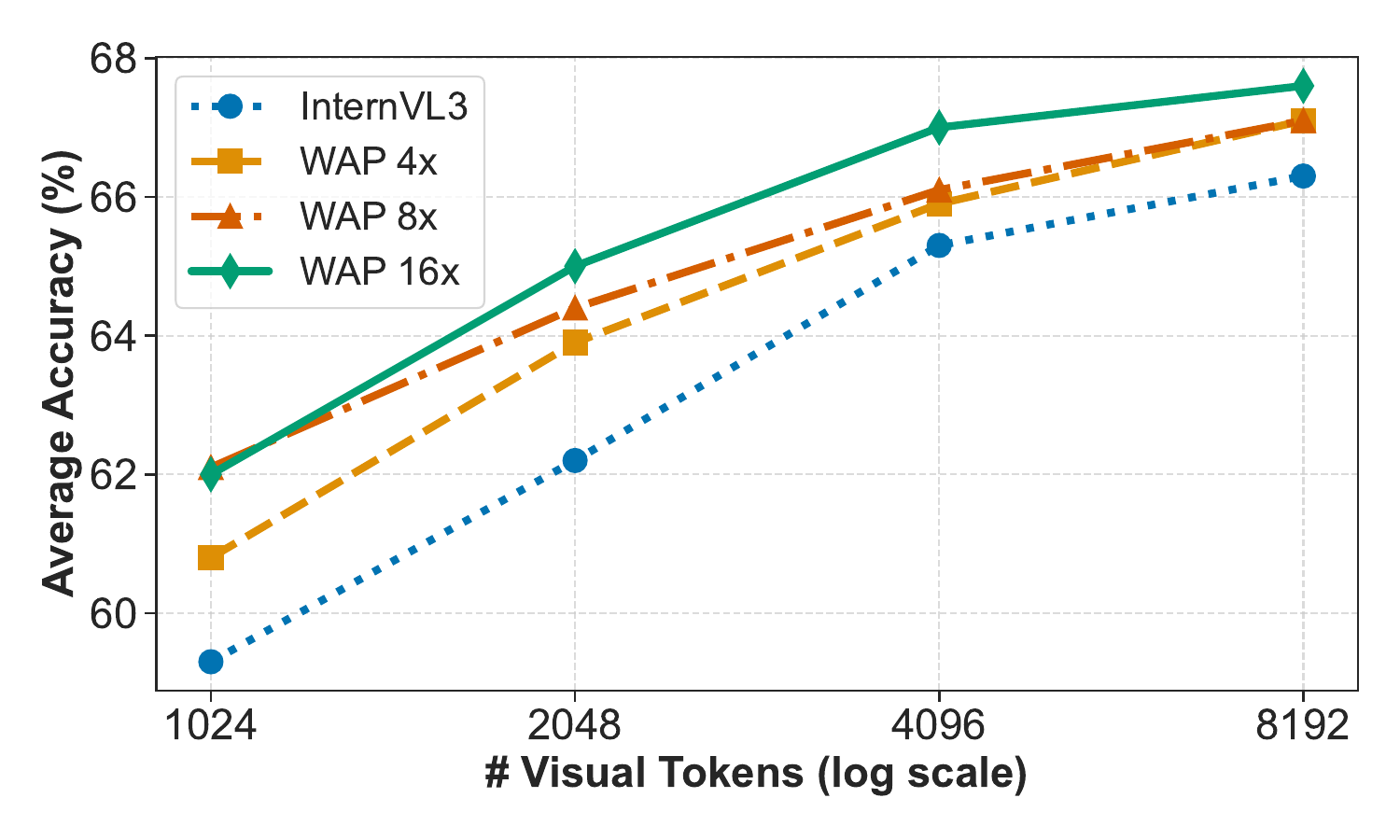} 
    \end{minipage}

    \caption{
        \textbf{Impact of frame scaling and test-time compression.} 
    \textbf{(Left)} Accuracy of Video LLMs on key benchmarks~(Video-MME, MLVU, and LongVideoBench) scales logarithmically with input frames, highlighting the strict frame budget as a primary bottleneck. 
        \textbf{(Right)} Under a fixed token budget, aggressive Weighted Average Pooling (up to 16$\times$, green) enables InternVL3-8B to process significantly more frames at test time, maximizing accuracy gains.
    }
    \label{fig:frame_scaling_extrapolation}
\end{figure}

\section{Revisiting Post-Hoc Reduction}
\label{sec:prelim}
In this section, we motivate the core design choices for LiteFrame~(\Cref{sec:method}).
We revisit post-hoc token reduction to establish two critical design premises for our main approach: (1) \textbf{W}eighted \textbf{A}verage \textbf{P}ooling~(WAP) serves as a simple and effective compression primitive compared to existing complex token merging or pruning strategies~(\Cref{subsec:wap}), and (2) aggressive compression~(up to $16\times$) is desirable because it trades off favorably with an increase in the number of frames that are processed at test time~(\Cref{subsec:extrapolation}).
Moreover, we demonstrate that post-hoc reduction fails to reduce the base computational cost of the encoder, prompting us to instead ``internalize'' the token compression via a customized compact student network architecture.

\begin{table}[t]
    \centering
    \caption{
        \textbf{Evaluation of post-hoc token reduction strategies.}
        We evaluate various token reduction methods applied to InternVL3-8B with a fixed compression ratio of $16\times$ (64 frames) on three video benchmarks~\citep{fu2025videomme,zhou2025mlvu,wu2024longvideobench}.
        The proposed \textbf{W}eighted \textbf{A}verage \textbf{P}ooling~(WAP) achieves the highest average accuracy.
        }
    \label{tab:wap_comparison}
    \setlength{\tabcolsep}{2pt}
    \scalebox{0.7}{
    \begin{tabular}{l|ccccc}
    \toprule
    \multirow{2}{*}{\textbf{Method}} & \multicolumn{2}{c}{\textbf{Video-MME}} & \multirow{2}{*}{\textbf{MLVU}} & \textbf{LongVideo} & \multirow{2}{*}{\textbf{Avg.}} \\
     & \textbf{(w/o subs)} & \textbf{(w subs)} &  & \textbf{Bench} & \\
    \midrule
    \rowcolor{gray!10} InternVL3-8B~\scriptsize{(No comp.)} & 66.0 & 68.7 & 73.1 & 59.6 & 66.9 \\
    \midrule
    Average Pooling & 59.7 & 64.1 & 62.3 & \textbf{54.7} & 60.2 \\
    Max Pooling & 59.7 & 63.9 & 62.0 & 54.2 & 60.0 \\
    Subsampling & 60.4 & 64.6 & 65.1 & 54.5 & 61.2 \\
    ToMe~\citep{bolya2023tome} & 58.7 & 62.3 & 64.7 & 54.2 & 60.0 \\
    PruMerge~\citep{shang2025llavaprumerge} & 60.3 & 63.2 & 64.6 & 50.6 & 59.7 \\
    FastVID~\citep{shen2025fastvid} & 59.3 & 63.4 & 65.1 & 54.4 & 60.6 \\
    \midrule
    \textbf{WAP~\scriptsize{(Ours)}} & \textbf{61.0} & \textbf{65.7} & \textbf{67.4} & 54.0 & \textbf{62.0} \\
    \bottomrule
    \end{tabular}%
    }
\end{table}

\subsection{Spatio-temporal Weighted Average Pooling (WAP)}
\label{subsec:wap}
To reduce the number of visual tokens, existing literature often relies on attention-based pruning~\citep{shang2025llavaprumerge,shen2025fastvid} or token merging via bipartite soft-matching~\citep{bolya2023tome,wang2025internvideo2}.
Since the attention and matching scores are mainly determined by the tokens' content rather than their positions, these methods disrupt the continuous spatio-temporal structure required for coherent video understanding.
Recent findings~\citep{wen2025token, liao2025vtcbench} highlight this drawback, suggesting that simple average pooling or image downsampling outperforms complex reduction strategies.
Extending this intuition, we propose Weighted Average Pooling (WAP), a primitive that harmonizes the structural regularity of pooling with attention-based weighting.

Let $\mathbf{X} \in \mathbb{R}^{T \times H \times W \times C}$ be the input feature tensor.
We partition $\mathbf{X}$ into non-overlapping spatio-temporal blocks $\Omega_{u,v,s}$ to match a target compressed resolution $(t, h, w)$.
The compressed token $\mathbf{Y}_{u,v,s}$, derived by WAP, is computed as:
\begin{equation}
\mathbf{Y}_{u,v,s} = \sum_{(\tau, i, j) \in \Omega_{u,v,s}} 
\text{softmax}\left( \frac{\mathbf{x}_{\tau,\text{cls}}^{\top} \mathbf{x}_{\tau, i, j}}{\sqrt{C}} \right) \mathbf{x}_{\tau, i, j},
\end{equation}
where the softmax is computed within each block $\Omega_{u,v,s}$, $\mathbf{x}_{\tau, i, j} = \mathbf{X}[\tau, i, j, :]$, and $\mathbf{x}_{\tau, \text{cls}}$ is the class token of the $\tau^\text{th}$ frame.
This operation effectively retains high-activation features while reducing the token count by a factor of $r = \frac{THW}{thw}$.

Empirically, \Cref{tab:wap_comparison} demonstrates that WAP significantly outperforms both standard pooling baselines~(Average/Max Pooling, Subsampling) and state-of-the-art, more complex token reduction methods~\citep{shen2025fastvid, shang2025llavaprumerge, bolya2023tome} under a $16\times$~($4\times$ spatial and $4\times$ temporal) compression ratio.
\Cref{sec:app_implementation} provides the evaluation setups for \Cref{tab:wap_comparison}.
While modern Video LLMs~\citep{li2025f16,li2024llava,wang2025internvideo2} typically rely on simple pooling or ToMe~\citep{bolya2023tome}, we instead use WAP as a compression operator, not merely for preprocessing, but to \emph{generate supervision targets for our distillation framework} in~\Cref{subsec:CTD}.

\subsection{Frame-Count Bottleneck}
\label{subsec:extrapolation}
The performance of Video LLMs depends critically on the number of input frames.
As shown in~\Cref{fig:frame_scaling_extrapolation}~(left), accuracy on the long video benchmarks, such as Video-MME~\citep{fu2025videomme}, MLVU~\citep{zhou2025mlvu}, and LongVideoBench~\citep{wu2024longvideobench}, exhibits logarithmic growth with respect to the input frame count.
However, conventional models like InternVL3 are practically capped at $\sim$64 input frames due to both the context length limits of the LLM and the large number of tokens per frame (e.g., 256).
We argue that this dense per-frame tokenization is excessive, and that spatio-temporal token compression can overcome these bottlenecks.

To validate this, we compare a baseline without compression against three WAP variants with compression ratios of $4\times$, $8\times$, and $16\times$ under a fixed visual token budget.
Crucially, WAP enables high compression ratios, thereby allowing the model to process proportionally more frames.
As seen in~\Cref{fig:frame_scaling_extrapolation}~(right), all WAP variants outperform the baseline, with $16\times$ compression (and thus $16\times$ more frames) achieving the best results.
These results demonstrate that aggressive compression effectively trades redundant tokens for richer temporal context.
\Cref{sec:app_implementation} describes the detailed experimental setup for \Cref{fig:frame_scaling_extrapolation}.

\vspace{-4mm}
\paragraph{Scaling paradox.}
While post-hoc reduction effectively reduces the number of visual tokens fed to LLMs, the computational cost of the vision encoder remains the same.
Therefore, as we scale the frame counts needed for high performance on long-form video understanding, the vision encoder latency explodes and becomes the new bottleneck~(\Cref{fig:stacked_plots_latency} (b)).
This insight drives the design of LiteFrame---we focus on achieving the aforementioned aggressive compression directly within the vision encoder, rather than as a post-hoc stage.


\begin{figure*}[t!]
    \centering
    \includegraphics[width=1.0\linewidth]{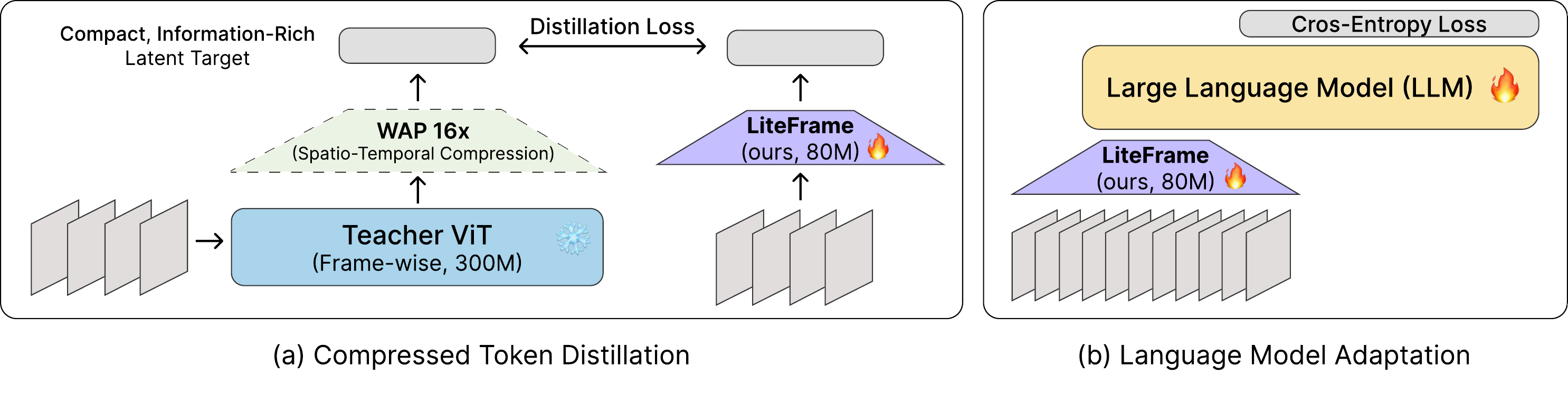}
    \vspace{-8mm}
    \caption{\textbf{Overview of our training framework for LiteFrame.} (a) Compressed Token Distillation employs WAP to compress the teacher's dense features into a compact, information-rich latent space, which serves as the prediction target for the student. (b) Language Model Adaptation fine-tunes the LLM and encoder on (video, text) pairs to further optimize the student's latent space and adapt LLM to the extended temporal context.}
\end{figure*}

\section{LiteFrame: Internalizing Spatio-Temporal Token Compression}
\label{sec:method}
We introduce LiteFrame, a video encoder designed to resolve the dual bottleneck of Video LLMs: the quadratic complexity of the LLM \textit{and} the exploding latency of the vision encoder when scaling to high input frame counts.
Unlike prior works that compress tokens post-hoc, we propose a lightweight encoder that internally compresses the tokens.
To achieve this, our approach rests on two key ideas.
First, we design a spatio-temporal encoder architecture that minimizes latency and FLOPs~(\Cref{subsec:architecture}).
Second, we propose a novel distillation strategy where the student learns to directly predict the spatio-temporally compressed representations of a powerful teacher~(\Cref{subsec:CTD}).

\subsection{Architecture: Spatio-temporal Token Compressive Encoding}
\label{subsec:architecture}

We first design a lightweight student encoder to be significantly more compact than the corresponding teacher~(87M vs.\ 304M parameters in our main experiments).
We use a 12-layer, 768D ViT-Base~\citep{dosovitskiy2021vit} backbone for the student while the teacher is a 24-layer, 1024D ViT-Large.

\begin{wraptable}{r}{0.5\linewidth}
\centering
\caption{\textbf{Efficiency comparison for temporal modeling.}
Compared to temporal attention~(TempAttn), spatio-temporal full attention~(SpatioTempAttn), or vanilla temporal convolutions~(TempConv),
Depth-Wise Temporal Convolutions~(DWTempConv) achieves the lowest latency and FLOPs while introducing negligible parameter overhead~($<$1M).
Latency and FLOPs are measured using 256 input frames.
}
\vspace{2mm}
\label{tab:arch_latency}
\setlength{\tabcolsep}{5pt}
\scalebox{0.74}{
\begin{tabular}{l c c c}
\toprule
\textbf{Architecture} & \textbf{Latency (ms)} & \textbf{TFLOPs} & \textbf{\# Params (M)} \\
\midrule
\rowcolor{gray!10}ViT-Large-24L~\scriptsize{(Teacher)} & 1043.33 & 158.80 & 304.01\\
\rowcolor{gray!10}ViT-Base-12L~\scriptsize{(No comp.)} & \ \ 338.01 & \ \ 44.81 & \ \ 86.31\\
\midrule
TempAttn & \ \ 348.29 & \ \ 32.77 & 143.83 \\
SpatioTempAttn & \ \ 204.35 & \ \ \textbf{17.92} &  \ \ \textbf{87.15} \\
TempConv & \ \ 202.08 & \ \ 22.44 & 109.54 \\
 \textbf{DWTempConv~\scriptsize{(Ours)}} & \ \ \textbf{174.84} & \ \ \textbf{17.92} & \ \ \textbf{87.15} \\
\bottomrule
\end{tabular}
}
\vspace{-2mm}
\end{wraptable}

Moreover, we employ a low-latency video encoder backbone---instead of the standard image encoder---designed to progressively reduce spatio-temporal redundancies across frames.
Specifically, to enable spatio-temporal encoding, we interleave standard spatial attention layers with lightweight, depth-wise~(DW) 1D temporal convolution layers.
To further reduce computation, we integrate DW strided convolution layers at strategic intervals, which gradually downsample the feature maps in both spatial and temporal dimensions as the network deepens.
By progressively reducing the number of tokens, we ensure that the computational cost of the deeper layers is substantially lower than that of standard frame-wise image encoders.
\Cref{subsec:implementation_details} describes the architecture in detail.

As demonstrated in \Cref{tab:arch_latency}, DW temporal convolutions allow the model to capture temporal dynamics with significantly lower latency and FLOPs, compared to other widely-used alternatives, such as interleaving temporal attention blocks, basic temporal convolution, or replacing the spatial attention with full spatio-temporal attention.
Moreover, \Cref{tab:ablation_study} demonstrates that DW temporal convolution consistently yields superior accuracy over full spatio-temporal attention across benchmarks.
\Cref{sec:app_implementation} details how the latency is measured.

\newcommand{\inc}[1]{\textcolor{teal}{\scriptsize{(#1)}}}   
\newcommand{\rem}[1]{\textcolor{gray}{\scriptsize{(#1)}}}
\newcommand{\dec}[1]{\textcolor{purple}{\scriptsize{(#1)}}} 
\newcommand{\neginc}[1]{\textcolor{purple}{\scriptsize{(#1)}}} 
\newcommand{\posdec}[1]{\textcolor{teal}{\scriptsize{(#1)}}} 

\vspace{-1mm}
\subsection{Compressed Token Distillation (CTD)}
\label{subsec:CTD}

Training a lightweight student to match the semantic richness of a large teacher while simultaneously reducing the token count is non-trivial.
Standard distillation forces the student to learn redundant spatial details that it cannot effectively represent.

To address this, we propose Compressed Token Distillation~(CTD), where we treat Weighted Average Pooling~(WAP) as a strong post-hoc compression primitive~(as seen in \Cref{subsec:wap}) and use it to generate supervision targets.
As a result, rather than mimicking the teacher's dense output, the student is trained to predict the compressed representation produced by the teacher under WAP.

Formally, let $T(x) = Z_T \in \mathbb{R}^{N \times D}$ denote the teacher's dense features and $S_\theta(x) = Z_S \in \mathbb{R}^{(N/r) \times D}$ denote the student's output, where $r$ is the target compression ratio (e.g., $16\times$).
We define a projection operator $\mathcal{P}(\cdot)$ based on WAP that aggregates dense tokens into compressed representations.
The student is optimized to minimize the MSE loss between its output and the teacher's compressed representations:
\begin{equation}
\mathcal{L}_{\text{CTD}}(\theta) = \| S_\theta(x) - \mathcal{P}(T(x)) \|_2^2.
\end{equation}
By effectively transferring the attention-based weighting mechanism of WAP into the static parameters of the student network, the student can output the salient spatio-temporal information without the runtime overhead of computing attention over redundant patches.

\subsection{Language Model Adaptation (LMA)}
\label{subsec:LA}
Although CTD effectively teaches the student to predict salient features, the resulting compressed latent space can be suboptimal for the LLM.
Therefore, to bridge the modality gap and further optimize the student's latent space, we add a minimal Language Model Adaptation~(LMA) stage.
We fine-tune the LLM and the encoder with video-text pairs, minimizing the standard cross-entropy loss for text generation conditioned on videos.
To ensure training efficiency and preserve the LLM's reasoning capabilities, we employ LoRA~\citep{hu2022lora}.
In addition to aligning the student with the LLM, we also find that this stage helps with \emph{long-context adaptation}, allowing the LLM to handle the extended temporal context (up to 512 frames) enabled by our encoder.

\begin{table*}[t!]
\centering
\caption{\textbf{Latency and accuracy trade-off.} 
We evaluate LiteFrame as an efficient video encoder, and FastVID~\citep{shen2025fastvid}, one of the state-of-the-art post-hoc methods, both applied to InternVL3-8B under comparable total latency budgets.
\textbf{LiteFrame~(Ours)} denotes our final performance incorporating Compressed Token Distillation and Language Model Adaptation. 
By internalizing token compression within the vision backbone, LiteFrame processes 8$\times$ more frames than the baseline while achieving up to a 35\% reduction in end-to-end latency and superior accuracy.
In contrast, the post-hoc method is bottlenecked by the heavy original vision encoder.}
\label{tab:main_results}

\setlength{\tabcolsep}{5pt}
\scalebox{0.72}{%
\begin{tabular}{llcccclccccl}
\toprule
\multirow{2}{*}{Method} & \multirow{2}{*}{Frames} & Tokens/ & Vision & \multicolumn{3}{c}{Latency (ms) $\downarrow$} & \multicolumn{5}{c}{Accuracy (\%) $\uparrow$} \\
\cmidrule(lr){5-7} \cmidrule(lr){8-12}
 & & Frame & Params & Vision & LLM & Total & V-MME$^{\dagger}$ & V-MME$^{\ddagger}$ & MLVU & LongVideo & Avg \\
\midrule
\rowcolor{gray!10} InternVL3-8B & 8 & 256 & 304M & 40.0 & 167.3 & 208.4 & 59.0 & 60.0 & 62.2 & \textbf{54.8}& 59.0 \\
+FastVID & 32 \scriptsize{}{(4$\times$)} & 16 & 304M & 161.7 & 63.0 & 224.8 \neginc{+7.9\%} & 58.7 & 62.5 & 64.0& 52.6 & 59.5 \inc{+0.5} \\
\textbf{+LiteFrame~\scriptsize{(Ours)}} & \textbf{64} \scriptsize{}{(8$\times$)} & 16 & \textbf{87M} & 54.8 & 94.9 & \textbf{150.1} \posdec{-28.0\%} & \textbf{61.0}& \textbf{64.2}& \textbf{65.7}& 53.6& \textbf{61.1} \inc{+2.1}\\
\midrule
\rowcolor{gray!10} InternVL3-8B & 16 & 256 & 304M & 74.0 & 329.3 & 403.6 & 61.9 & 64.0 & 66.4 & 56.5& 62.2 \\
+FastVID & 64 \scriptsize{}{(4$\times$)} & 16 & 304M & 310.6 & 95.4 & 406.2 \neginc{+0.6\%} & 59.3 & 63.4 & 65.1 & 52.6 & 59.5 \dec{-2.7} \\
\textbf{+LiteFrame~\scriptsize{(Ours)}} & \textbf{128} \scriptsize{}{(8$\times$)} & 16 & \textbf{87M} & 105.3 & 166.6 & \textbf{272.6} \posdec{-32.5\%} & \textbf{63.9}& \textbf{66.8}& \textbf{66.7}& \textbf{57.2}& \textbf{63.7} \inc{+1.5}\\
\midrule
\rowcolor{gray!10} InternVL3-8B & 32 & 256 & 304M & 144.5 & 669.8 & 814.5 & \textbf{65.6} & 67.4 & 69.6 & \textbf{58.7} & 65.3 \\
+FastVID & 128 \scriptsize{(4$\times$)} & 16 & 304M & 625.8 & 168.9 & 794.9 \posdec{-2.4\%} & 60.4 & 66.1 & 69.3 & 55.6 & 62.9 \dec{-2.4} \\
\textbf{+LiteFrame~\scriptsize{(Ours)}} & \textbf{256} \scriptsize{}{(8$\times$)} & 16 & \textbf{87M} & 204.0 & 327.4 & \textbf{532.3} \posdec{-34.6\%} & 65.1& \textbf{68.5}& \textbf{70.7}& 58.3& \textbf{65.7} \inc{+0.4}\\
\bottomrule
\end{tabular}
}
\\ \footnotesize{$^{\dagger}$ without subtitles, $^{\ddagger}$ with subtitles. \inc{+}/\dec{-} denotes improvement/degradation relative to Teacher.}
\end{table*}

\section{Experiments}
\label{sec:exp}

\subsection{Implementation details}
\label{subsec:implementation_details}

We utilize InternVL3-8B as our primary baseline, leveraging its image encoder, InternViT-300M~(304M parameters, 1024 hidden dim), as the teacher model.
To measure the average accuracy, we employ four widely used video benchmarks---Video-MME~(with and without subtitiles; \citealp{fu2025videomme}), MLVU~\citep{zhou2025mlvu}, and LongVideoBench~\citep{wu2024longvideobench}---as primary evaluation suites.

For the student model, we adopt a significantly more efficient ViT-Base backbone~(87M parameters, 768 hidden dimensions).
As described in \Cref{subsec:architecture}, we interleave depth-wise 1D temporal convolutions after every spatial layer where the temporal dimension is greater than 1.
In addition, we integrate depth-wise strided convolution layers after the 4$^\text{th}$ and 8$^\text{th}$ blocks, with strides of $[t,h,w]=[2,2,2]$ and $[2,1,1]$, respectively.
Further details regarding training, datasets, and evaluation are provided in \Cref{sec:app_implementation}.

\subsection{Quantitative analysis}
\subsubsection{Redefining the Pareto frontier}
We evaluate LiteFrame by analyzing the trade-off between video understanding accuracy across multiple benchmarks~\citep{fu2025videomme,zhou2025mlvu,wu2024longvideobench} and end-to-end inference latency under varying frame counts.
As detailed in \Cref{tab:main_results}, our approach establishes a new Pareto frontier, surpassing the baselines in both latency and accuracy. 
Applied to InternVL3-8B, LiteFrame reduces total inference latency by up to 35\% while improving accuracy by 0.4\%p~(65.7\% vs.\ 65.3\%) on average.
Notably, the accuracy gap widens by 2.1\%p~(61.1\% vs.\ 59.0\%), when we restrict the total latency budget~(8 frames for InternVL3-8B).
Moreover, as shown in \Cref{fig:teaser}, LiteFrame significantly outperforms state-of-the-art post-hoc compression methods such as FastVID~\citep{chen2024fastv}, PruMerge~\citep{shang2025llavaprumerge}, and ToMe~\citep{bolya2023tome}.
The results demonstrate that LiteFrame effectively trades spatio-temporal redundancy for significantly richer temporal context, allowing the model to process $8\times$ more frames within a fixed compute budget.

\begin{figure}[!t]
    \centering
    \begin{minipage}[t]{0.48\textwidth}
        \centering
        \includegraphics[width=\linewidth]{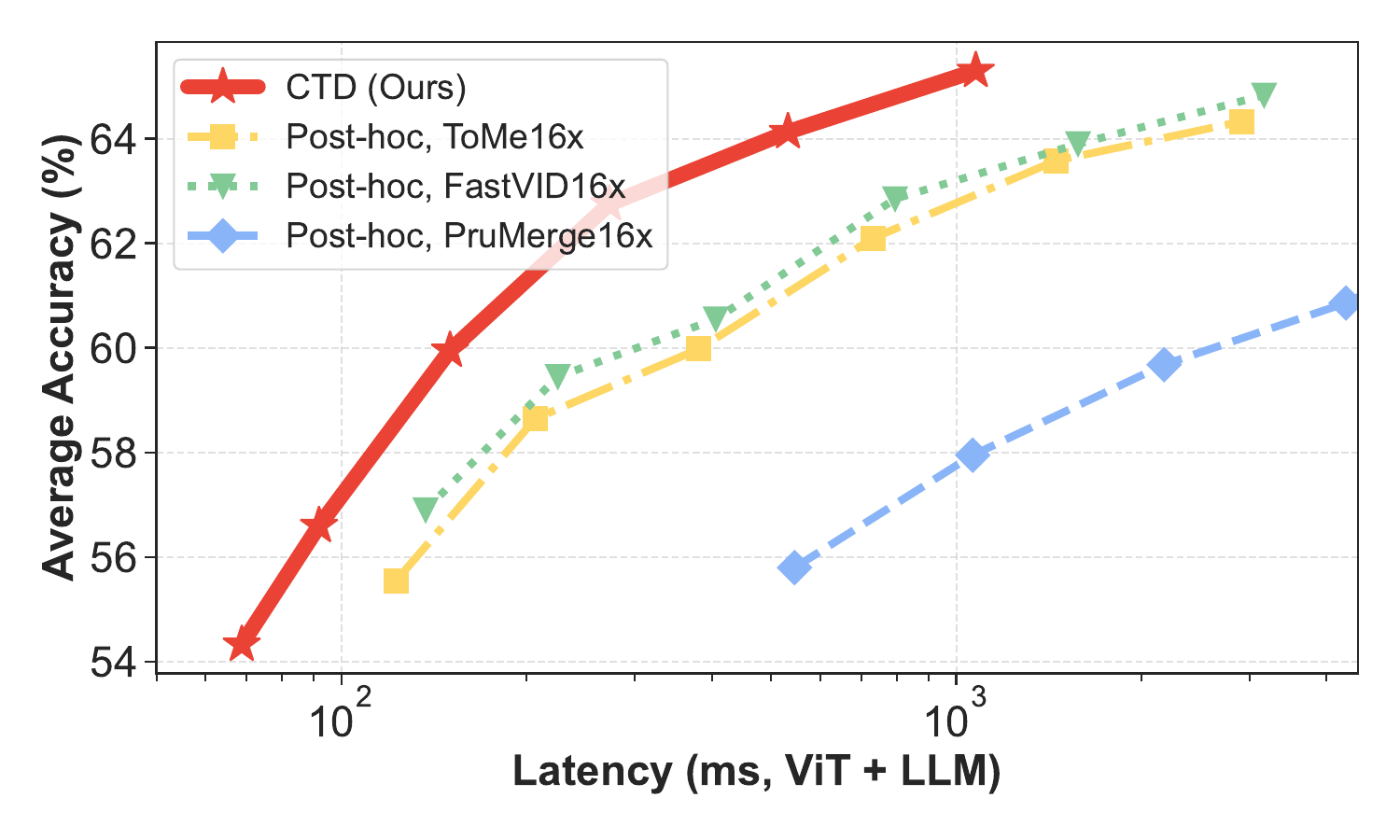} \vspace{-6mm}
        \caption{\textbf{Comparison with SOTA post-hoc token compression methods.}
        Compressed Token Distillation~(CTD) achieves superior accuracy compared to the baselines by avoiding the computational floor that limits the efficiency.}
        \label{fig:comp_posthoc}
    \end{minipage}\hfill
    \begin{minipage}[t]{0.48\textwidth}
        \centering
        \includegraphics[width=\linewidth]{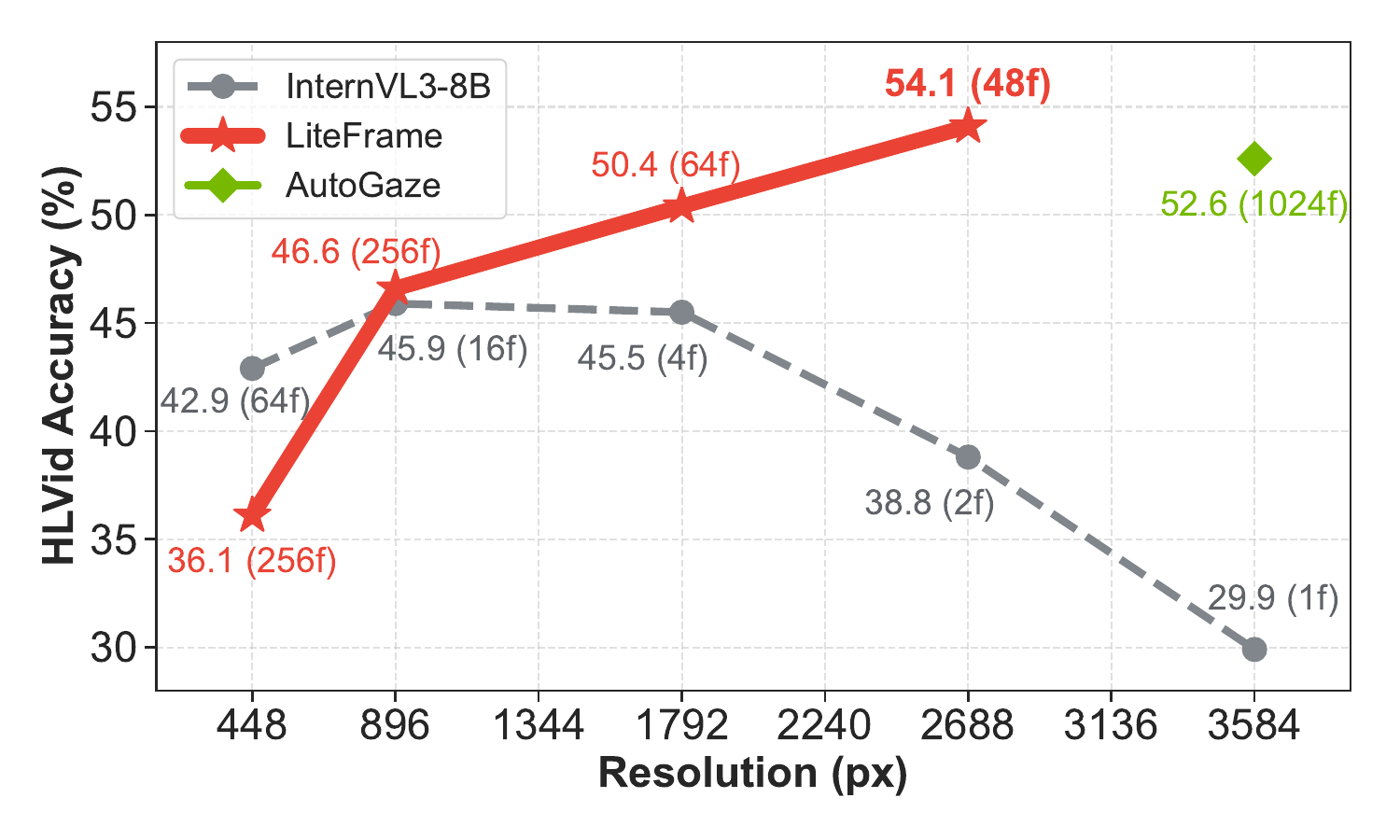} \vspace{-6mm}
        \caption{\textbf{Zero-shot scaling in spatial dimension.}
        We measure maximum accuracy across spatial resolutions, utilizing the maximum context length of the LLMs.
        LiteFrame achieves a state-of-the-art score of \textbf{54.1} on HLVid.}
        \label{fig:comp_spatial_resolution}
    \end{minipage}
\end{figure}

\subsubsection{Comparison with post-hoc methods}
To ensure a fair comparison with training-free post-hoc baselines, we evaluate LiteFrame utilizing only CTD without subsequent LMA, keeping the LLM entirely frozen.
As illustrated in \Cref{fig:comp_posthoc}, simply swapping the original heavy ViT with LiteFrame surpasses all post-hoc methods---including ToMe~\citep{bolya2023tome}, LLaVA-PruMerge~\citep{shang2025llavaprumerge}, and FastVID~\citep{shen2025fastvid}---in both efficiency and accuracy, by effectively distilling the WAP primitive into the student model.
In contrast, as expected, existing post-hoc methods are severely bottlenecked by the inevitable computational cost incurred prior to the compression, causing ViT latency to explode when frame counts increase.

\subsubsection{Zero-shot spatial resolution scaling}
\label{subsubsec:spatial_scaling}
Beyond scaling temporal resolution for long-form video, the inherent token efficiency of LiteFrame naturally facilitates scaling in the spatial dimension, particularly for tasks requiring fine-grained visual perception.
To highlight this, we implement a zero-shot tiling strategy that splits high-resolution frames into 448px sub-tiled clips, which are then processed independently by LiteFrame.
We evaluate this on the HLVid benchmark~\citep{shi2026autogaze}, which requires high-fidelity spatial understanding across video frames~(see \Cref{fig:comp_spatial_resolution}).
Notably, InternVL3-8B exhibits a performance stagnation as input resolution increases---we attribute this to the LLM's fixed context length that forces a sacrifice in temporal resolution as token counts grow due to the increased spatial resolution.
In contrast, the token efficiency of LiteFrame allows the model to maintain a better balance of spatial and temporal resolution, achieving a state-of-the-art score of 54.1 at 2688px with 48 frames.
Remarkably, this surpasses the previous best method, AutoGaze~\citep{shi2026autogaze}~(52.6), despite AutoGaze requiring much higher resolutions~(3584px and 1024 frames).
Moreover, LiteFrame \textit{achieves these results without any high-resolution training}, demonstrating its strong generalizability to higher resolutions.

\begin{table*}[t]
    \centering
    \vspace{-2mm}
    \caption{\textbf{Comparison with efficient vision encoders for MLLMs.} We compare LiteFrame against state-of-the-art efficient vision encoders for VLMs.
    Our method shows significantly lower latency for both vision encoding and LLM prefilling, while achieving the best accuracy.
    ``Total Tok.'' denotes the aggregate number of visual tokens fed to the LLM, and ``Acc.'' denotes the average accuracy. 
    }
    \vspace{2mm}
    \setlength{\tabcolsep}{8pt}
    \scalebox{0.7}{%
    \begin{tabular}{lcccccccc}
        \toprule
        \textbf{Method} & \textbf{Vision Params} & \textbf{Tok./Frame} & \textbf{Frames} & \textbf{Total Tok.} & \textbf{Vis. (ms)} & \textbf{LLM (ms)} & \textbf{Total (ms)} & \textbf{Acc. (\%)} \\
        \midrule
        FastVLM & 125M & 49 & 32 & 1568 & 98.3 & 132.9 & 231.5 & 47.6 \\
        VideoPanda & 45M & 272 & 32 & 8704 & 36.5 & 345.7 & 382.4 & 49.2 \\
        \textbf{LiteFrame~\scriptsize{(Ours)}} & 87M & \textbf{16} & 32 & \textbf{512} & \textbf{30.1} & \textbf{61.5} & \textbf{91.9} & \textbf{58.0} \\
        \bottomrule
    \end{tabular}
    }
    \label{tab:sota_efficiency}
\end{table*}

\subsubsection{Comparison with efficient vision encoders for MLLMs}
We further benchmark our approach against state-of-the-art efficient vision encoders designed for MLLMs, including FastVLM~\citep{vasu2025fastvlm} and VideoPanda~\citep{yi2025videopanda}.
Based on InternVL3-8B, we fine-tune the LLM via LoRA with the respective frozen visual encoders for a fair comparison.
As shown in \Cref{tab:sota_efficiency}, while these baselines achieve impressive parameter efficiency, LiteFrame is $1.2\times$ faster than VideoPanda and $3.3\times$ faster than FastVLM.
Additionally, because our encoder is trained to output a highly compact set of tokens, it also significantly lowers the downstream computational cost~(and thus latency) of the LLM.

\begin{figure}[!t]
    \centering
    \begin{minipage}[c]{0.34\linewidth}
        \centering
        \vspace{2mm}
        \includegraphics[width=\linewidth]{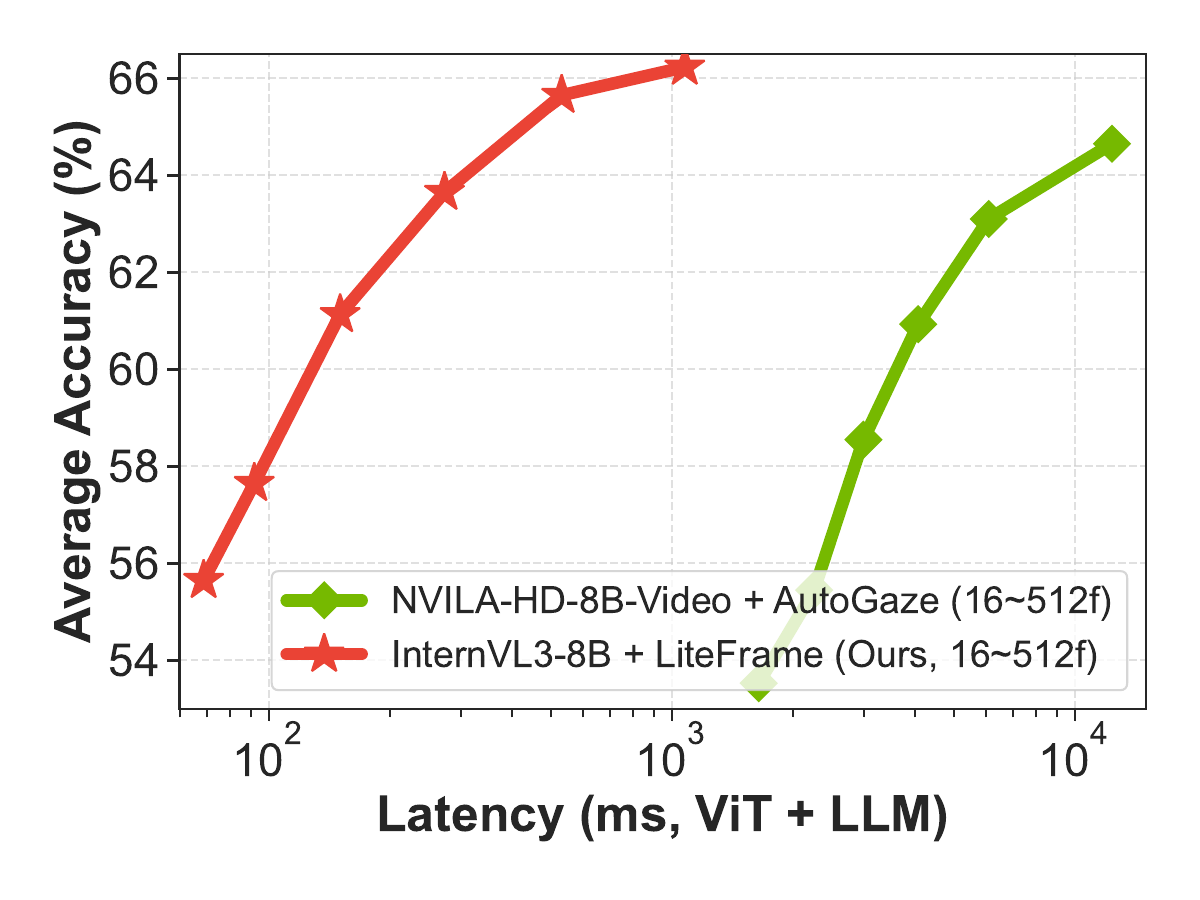}
    \end{minipage}\hfill
    \begin{minipage}[c]{0.62\linewidth}
        \vspace{-4mm}
        \setlength{\tabcolsep}{6pt}
        \centering
        \scalebox{0.7}{%
        \begin{tabular}{lcccccc}
        \toprule
        \multirow{2}{*}{Method} & \multirow{2}{*}{Frames} & \multicolumn{4}{c}{Latency (ms)} & \multirow{2}{*}{Acc. (\%)} \\
        \cmidrule(lr){3-6}
         & & AutoGaze & ViT & LLM & Total & \\
        \midrule
        \rowcolor{gray!10} NVILA-8B-Video & 32 & - & 451.0 & 329.9 & 780.8 & 63.1 \\
        +AutoGaze & \begin{tabular}{@{}c@{}}256 \\ \scriptsize{}{(8$\times$)}\end{tabular} & \begin{tabular}{@{}c@{}}2961.4 \\ \neginc{+2961.4ms}\end{tabular} & \begin{tabular}{@{}c@{}}2605.6 \\ \neginc{+477.8\%}\end{tabular} & \begin{tabular}{@{}c@{}}539.5 \\ \neginc{+63.5\%}\end{tabular} & \begin{tabular}{@{}c@{}}6106.5 \\ \neginc{+682.1\%}\end{tabular} & \begin{tabular}{@{}c@{}}63.1 \\ \rem{0.00}\end{tabular} \\
        \midrule
        \rowcolor{gray!10} InternVL3-8B & 32 & - & 144.5 & 669.8 & 814.5 & 65.3 \\
        +LiteFrame~\scriptsize{(Ours)} & \begin{tabular}{@{}c@{}}256 \\ \scriptsize{}{(8$\times$)}\end{tabular} & - & \begin{tabular}{@{}c@{}}{204.0} \\ \neginc{+41.2\%}\end{tabular} & \begin{tabular}{@{}c@{}}{327.4} \\ \posdec{-51.1\%}\end{tabular} & \begin{tabular}{@{}c@{}}\textbf{532.3} \\ \posdec{-34.6\%}\end{tabular} & \begin{tabular}{@{}c@{}}\textbf{65.7} \\ \inc{+0.4}\end{tabular} \\
        \bottomrule
        \end{tabular}
        }
    \end{minipage}


    \caption{
        \textbf{Comparison with AutoGaze.} 
        \textbf{(Left)} LiteFrame~(red) improves the efficiency frontier with lower latency and higher accuracy compared to AutoGaze~(green).
        \textbf{(Right)} When scaling from 32 to 256 frames ($8\times$), LiteFrame balances ViT and LLM scaling, while AutoGaze's pre-reduction module becomes a new bottleneck.
        ``Acc.'' denotes the average accuracy.
    }
    \label{fig:movinet_autogaze_combined}
\end{figure}

Next, we compare LiteFrame against AutoGaze~\citep{shi2026autogaze}, a recent approach that similarly addresses the computational bottlenecks of both the ViT and LLM.
We benchmark the latency-accuracy trade-offs when integrating LiteFrame and AutoGaze into their respective baselines.
As shown in \Cref{fig:movinet_autogaze_combined}~(left), LiteFrame significantly outperforms AutoGaze,
achieving substantially lower total latency while improving in average accuracy.
While initially surprising, the detailed breakdown in \Cref{fig:movinet_autogaze_combined}~(right) reveals that the latency gap can be attributed to the AutoGaze pre-reduction auxiliary module that accounts for nearly half of the total inference time~(3.0s out of 6.1s).
A more detailed comparison and implementation details regarding AutoGaze are provided in \Cref{sec:app_autogaze}.

\begin{table}
    \centering
    \caption{\textbf{Ablation study.} We evaluate the impact of the token-compressive student architecture~(TokComp.), Depth-Wise Temporal Convolutions (DWConv), the Weighted Average Pooling (WAP) objective, and Language Model Adaptation (LMA).
    ``Acc'' denotes the average accuracy.
    Our approach~(bottom row) demonstrates the best trade-off, simultaneously reducing latency while improving accuracy.}
    \label{tab:ablation_study}
    \setlength{\tabcolsep}{2pt}
    \scalebox{0.75}{
    \begin{tabular}{l c c c c c c c}
    \toprule
    \multirow{2}{*}{\textbf{Ablations}} & \textbf{TokComp.} & \textbf{DWConv} & \textbf{WAP} & \multirow{2}{*}{\textbf{LMA}} & \multirow{2}{*}{\textbf{Frames}} & \textbf{Latency} & \textbf{Acc} \\
     & \textbf{(Student)} & \textbf{(Student)} & \textbf{(Teacher)} & & & \textbf{(ms)} & \textbf{(\%)} \\
    \midrule
    \rowcolor{gray!10} InternVL3-8B \scriptsize{(Teacher)} & -- & -- & -- & -- & \ \ 16 & 403.6 & 62.2 \\
    \midrule
    Distillation~\scriptsize{(ViT-Base-12L)} & $\times$ & $\times$ & $\times$ & $\times$ & \ \ 16 & 362.9 & 60.3 \\
    CTD~\scriptsize{(SpatioTempAttn)} & \checkmark & $\times$ & \checkmark & $\times$ & 128 & 102.2 & 61.9 \\
    CTD~\scriptsize{(DWTempConv)} & \checkmark & \checkmark & \checkmark & $\times$ & 128 & \ \ 87.4 & 62.8 \\
    RTD & \checkmark & \checkmark & $\times$ & $\times$ & 128 & \ \ 87.4 & 43.8 \\
    RTD + LMA & \checkmark & \checkmark & $\times$ & \checkmark & 128 & \ \ 87.4 & 61.5 \\
    \textbf{CTD + LMA~\scriptsize{(Ours)}} & \checkmark & \checkmark & \checkmark & \checkmark & 128 & \ \ \textbf{87.4} & \textbf{63.4} \\
    \bottomrule
    \end{tabular}
    }
\end{table}

\subsection{Ablation studies}
\label{subsec:ablation}
To evaluate the effectiveness of each component of our approach, we conduct an ablation analysis isolating the contributions of 1) the token-compressive student architecture, 2) depth-wise~(DW) temporal convolutions, 3) the Weighted Average Pooling~(WAP) objective applied to the teacher's output, and 4) Language Model Adaptation~(LMA).

Simple distillation into a standard ViT-Base-12L backbone without token compression shows marginal latency reduction and degrades accuracy, most likely due to the context limits, which restrict temporal resolution.
Incorporating Compressed Token Distillation~(CTD) significantly alleviates this bottleneck, however, utilizing full spatio-temporal attention in the student encoder underperforms DW temporal convolutions, highlighting the superior efficiency and efficacy of this form of temporal processing.
Moreover, we explore an alternative training objective, Reconstructive Token Distillation~(RTD), which replaces our WAP objective with an auto-encoding objective~(detailed in \Cref{subsec:ablation_training}).
RTD significantly lags behind CTD, demonstrating that the WAP primitive enables more effective distillation of strong spatio-temporal features into the student for downstream LLM reasoning.
Finally, coupling CTD with the additional LMA stage ultimately gives the lowest latency and best accuracy.
A more comprehensive ablation analysis is provided in~\Cref{sec:app_ablation}.

\section{Conclusion}
\label{sec:conclusion}
In this work, we identify and resolve a critical efficiency bottleneck in current Video LLMs.
While post-hoc token reduction strategies effectively reduce the computational cost of the LLM, this leaves the vision encoder as the prohibitive latency bottleneck when scaling to high frame counts.
We introduce LiteFrame, a lightweight video encoder that fundamentally addresses the full end-to-end efficiency problem by internalizing spatio-temporal compression in a compact encoder, trained with our novel Compressed Token Distillation~(CTD) and Language Model Adaptation~(LMA) methods.
By teaching the student encoder to bypass redundant full-resolution computation and directly predict the information-dense~(pooled) tokens of the heavy teacher, our approach redefines the efficiency-accuracy Pareto frontier.
Specifically, across multiple long-video benchmarks, we achieve 35\% faster end-to-end inference and better accuracy while processing $8\times$ more frames, effectively trading spatio-temporal redundancy for significantly richer temporal context.
While much of the community has been focused exclusively on pushing the limits of pure token reduction methods, our results demonstrate the principle that architectural internalization of token compression via distillation unlocks even more scalable, long-form video understanding.

\newpage
\bibliography{main}
\newpage

\appendix

\section{Implementation details}
\label{sec:app_implementation}
\paragraph{Training~(CTD).}
For Compressed Token Distillation~(CTD), we train the student encoder using the AdamW optimizer with a cosine learning rate schedule and linear warmup.
Before distillation, we initialize the student's weights from those of the teacher, clipping them to match the student's dimensions.
The global batch size is set to 512, distributed across 8$\times$ NVIDIA H100 GPUs. The maximum learning rate is set to 4e-5 with a 100 epochs warmup period;
for specific variants susceptible to training instability, we reduce the learning rate to prevent loss explosion.
The total training duration is 1800 epochs, requiring approximately 21 days.
For the ablation studies~(\Cref{tab:ablation_study,tab:ablation_variants,tab:ablation_strategies}), we perform only 800 epochs of distillation for the training efficiency.
We sample 4-frame clips with a frame rate (FPS) uniformly sampled from $[1, 4]$.
To stabilize training, we apply an MSE outlier clipping strategy, clipping the target-prediction differences that exceed $3\times$ the standard deviation.
Furthermore, we employ gradient clipping with a maximum norm of 1.0.

\paragraph{Training~(LMA).}
For Language Model Adaptation~(LMA), we adapt the LLM using Low-Rank Adaptation~\citep{hu2022lora} with rank $r=4$, $\alpha=8$, and $\text{lora\_dropout}=0.05$.
Extensive experiments demonstrate that a lower rank~(e.g.\ 4) performs better than higher ones~(e.g.\ 8 and 16).
We employ an effective batch size of 128 using gradient accumulation and train with a learning rate of 4e-5 following a cosine schedule.
During this phase, we uniformly sample frame counts from $\{128, 256, 512\}$ with an FPS ranging from 1 to 4.
This ensures that the total visual token volume matches that of the teacher's typical input (equivalent to 8–32 frames for the uncompressed teacher).
We perform LMA on 8$\times$ NVIDIA H100 GPUs for 25K steps, which completes in a few hours.

\paragraph{Datasets.}
Our training pipeline utilizes a subset of the video data described in InternVL2.5 paper~\citep{chen2024internvl25}.
To be specific, we employ a diverse mix of datasets including ShareGPT4Video \citep{chen2024sharegpt4video}, LLaVA-Video-178K~\citep{zhang2024llavavideo}, FineVideo~\citep{xu2024finevideo}, CLEVRER~\citep{yi2020clevrer}, and NTURGB+D~\citep{shahroudy2016ntu} for CTD.
Moreover, we employ high-quality video-question answering pairs from LLaVA-Video-178K and FineVideo, alongside captioning datasets from ShareGPT4Video and OpenVid-1M~\citep{nan2024openvid1m} to ensure robust visual-textual alignment.
The datasets used in our work adhere to their respective license: ShareGPT4Video (CC-BY-NC-4.0), FineVideo (CC-BY), OpenVid-1M (CC-BY 4.0), and LLaVA-Video-178K~(Apache License 2.0). Note that CLEVRER, and NTURGB+D are exclusively restricted to non-commercial, academic research purposes.

\paragraph{Benchmarks.}
We employ three widely used video benchmarks---Video-MME~\citep{fu2025videomme}, MLVU~\citep{zhou2025mlvu}, and LongVideoBench~\citep{wu2024longvideobench}---as primary evaluation suites, measuring the average performance.
Additionally, HLVid~\citep{shi2026autogaze} is employed to evaluate high-fidelity spatial understanding capabilities for \Cref{fig:comp_spatial_resolution}.
Furthermore, we report the latency-accuracy trade-offs on short video benchmarks, such as MVBench~\citep{li2024mvbench} and TVbench~\citep{cores2024tvbench}, as well as additional long video benchmarks, including LVBench~\citep{wang2024lvbench} and MMBench-Video~\citep{fang2024mmbenchvideo}, in \Cref{sec:app_more_benchmarks}.
The datasets evaluated in this work strictly adhere to their respective licenses: MVBench (MIT), HLVid (Apache 2.0), TVBench and MMBench-Video (CC-BY-4.0), and LongVideoBench, MLVU, and LVBench (CC-BY-NC-SA-4.0).
Note that Video-MME is exclusively restricted to non-commercial, academic research purposes.

\paragraph{Evaluation setups.}
For evaluating InternVL3-8B~\citep{zhu2025internvl3} and the post-hoc methods in \Cref{fig:frame_scaling_extrapolation,tab:wap_comparison,fig:comp_posthoc}, we uniformly sample frames across the entire video and resize them to 448px.
The three WAP variants---WAP 4$\times$, 8$\times$, and 16$\times$---in \Cref{fig:frame_scaling_extrapolation} (right) employ compression ratios of $(t,h,w)=(1,2,2)$, $(2,2,2)$, and $(4,2,2)$, respectively.
For evaluating LiteFrame, we adopt a dense clip sampling strategy, unlike standard uniform frame sampling.
We uniformly sample multiple clips across the video, where each clip consists of a fixed number of frames~(4) extracted at a minimum of 1 FPS.
All sampled frames are resized to 448px before being fed into the encoder to match the original evaluation setup of the baseline model, except for the spatial scaling experiments presented in \Cref{fig:comp_spatial_resolution}.

\paragraph{Latency.}
Latency is measured end-to-end including ViT processing and LLM prefilling.
We focus exclusively on the visual token encoding and its prefilling stage, as these constitute the primary bottleneck addressed by our contributions.
We report the median latency over 100 iterations, following a 40 iterations of warmup phase (140 iterations total), measured on a single NVIDIA A100-80GB GPU.

\section{Additional video benchmarks}
\label{sec:app_more_benchmarks}

\subsection{Short video benchmarks}
The efficacy of LiteFrame extends beyond long-form video understanding, demonstrating natural applicability to short video tasks.
Specifically, LiteFrame reduces end-to-end latency by 28\% and 63\% on MVBench~\citep{li2024mvbench} and TVBench~\citep{cores2024tvbench}, respectively, while maintaining the accuracy of the baseline.

\begin{figure}[!h]
    \centering
    \begin{minipage}[b]{0.48\textwidth}
        \centering
        \includegraphics[width=\linewidth]{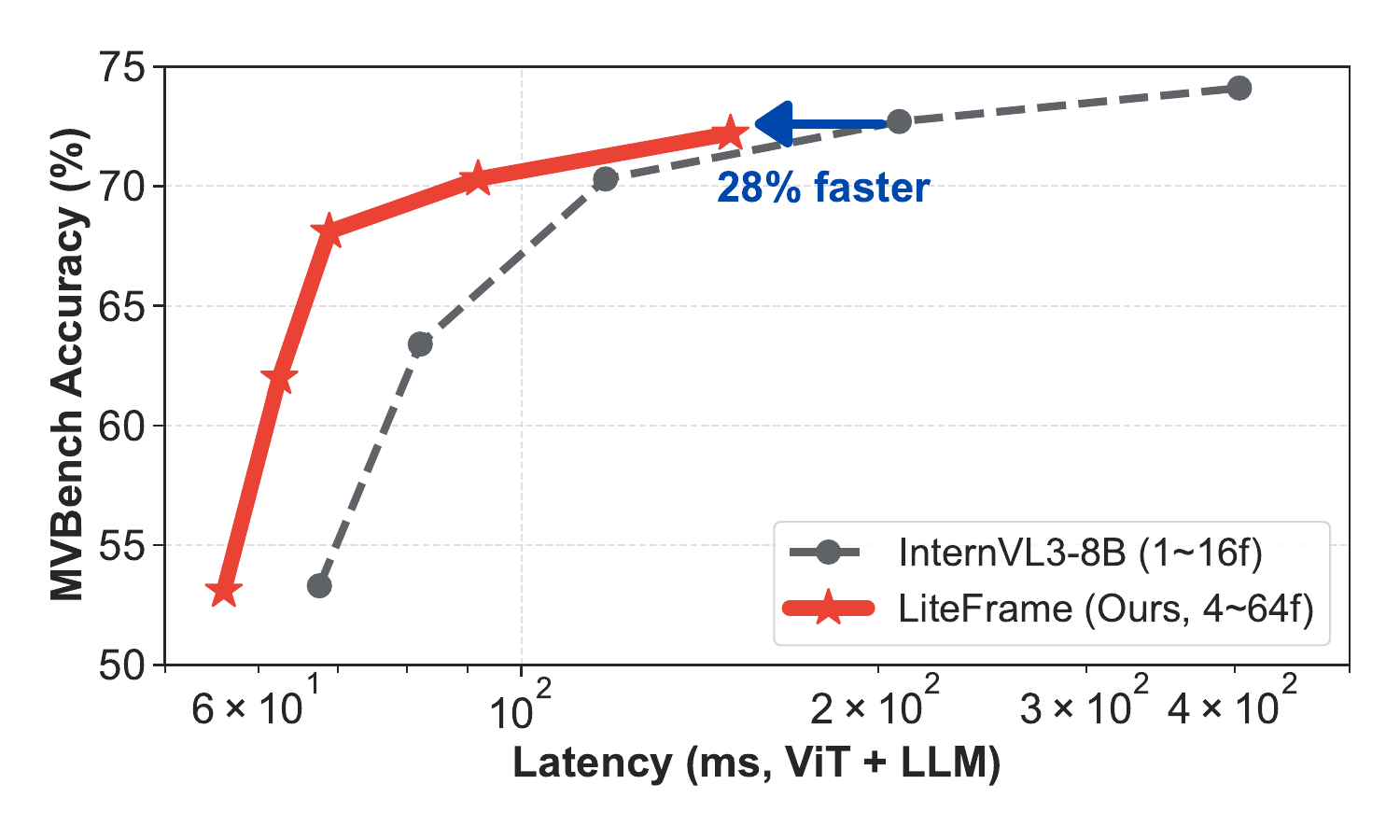} 
    \end{minipage}\hfill
    \begin{minipage}[b]{0.48\textwidth}
        \centering
        \includegraphics[width=\linewidth]{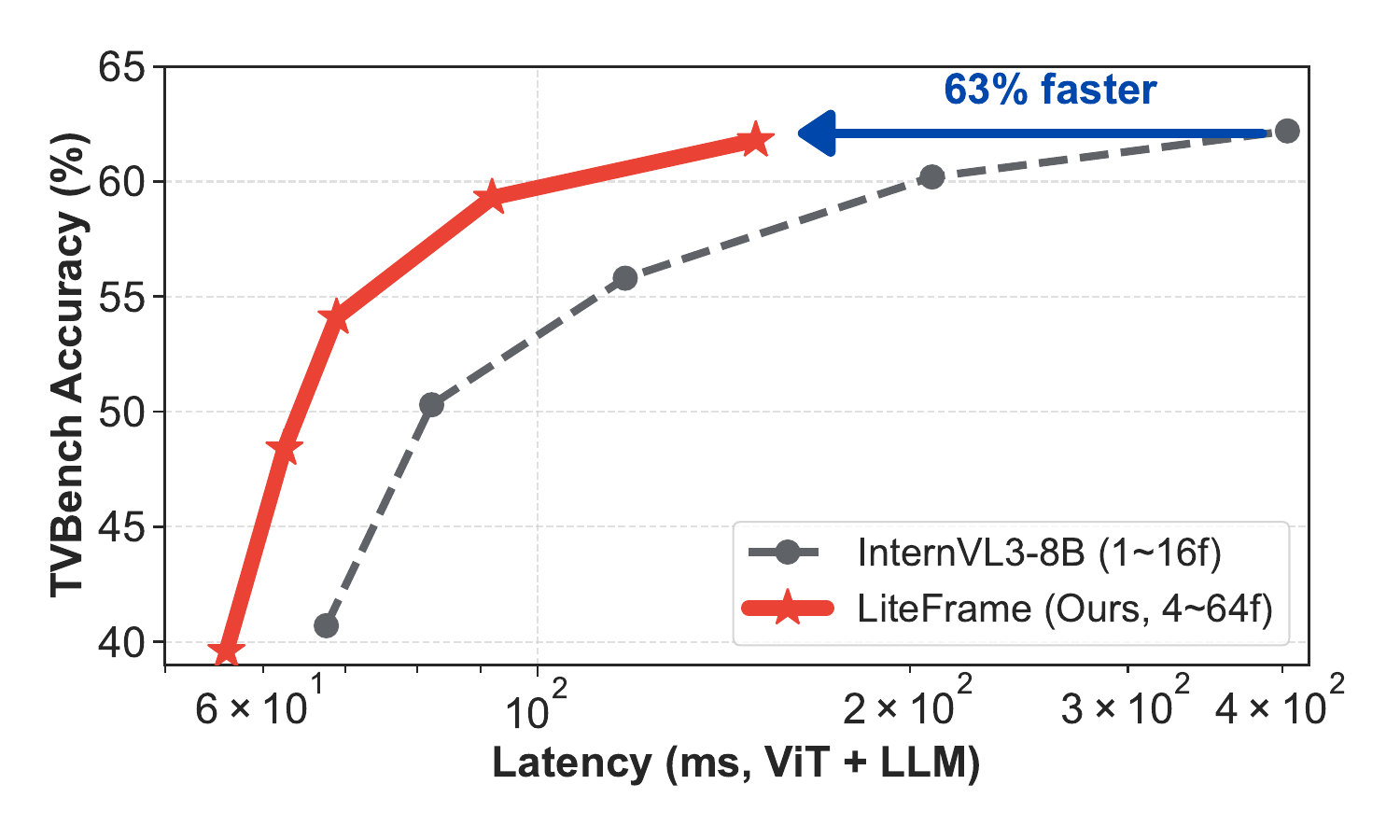} 
    \end{minipage}
    
    \vspace{2mm} 

    \caption{
        \textbf{Latency-accuracy trade-offs on short video benchmarks.}
        Evaluation on MVBench~\textbf{(Left)} and TVBench~\textbf{(Right)}.
        Even across standard short-form evaluation setups, LiteFrame achieves significant latency reductions while preserving accuracy.
    }
    \label{fig:mvbench_tvbench}
\end{figure}

\subsection{Long video benchmarks}
\label{subsec:app_long_video_benchmarks}
We report additional results on two long video benchmarks, LVBench~\citep{wang2024lvbench} and MMBench-Video~\citep{fang2024mmbenchvideo}.
Notably, on LVBench, LiteFrame utilizing 512-frame input achieves a superior score of 43.9 compared to the 64-frame baseline~(43.5) while operating 38\% faster, successfully leveraging the extended temporal context.
On MMBench-Video, a free-form QA benchmark, LiteFrame demonstrates improved efficiency, particularly within the low-latency regime~(16--128 input frames).
To evaluate the response quality on the MMBench-Video, we employ the Gemini 3 Flash Preview API~\citep{google2025gemini3flashpreview}.

\begin{figure}[!h]
    \centering
    \begin{minipage}[b]{0.48\textwidth}
        \centering
        \includegraphics[width=\linewidth]{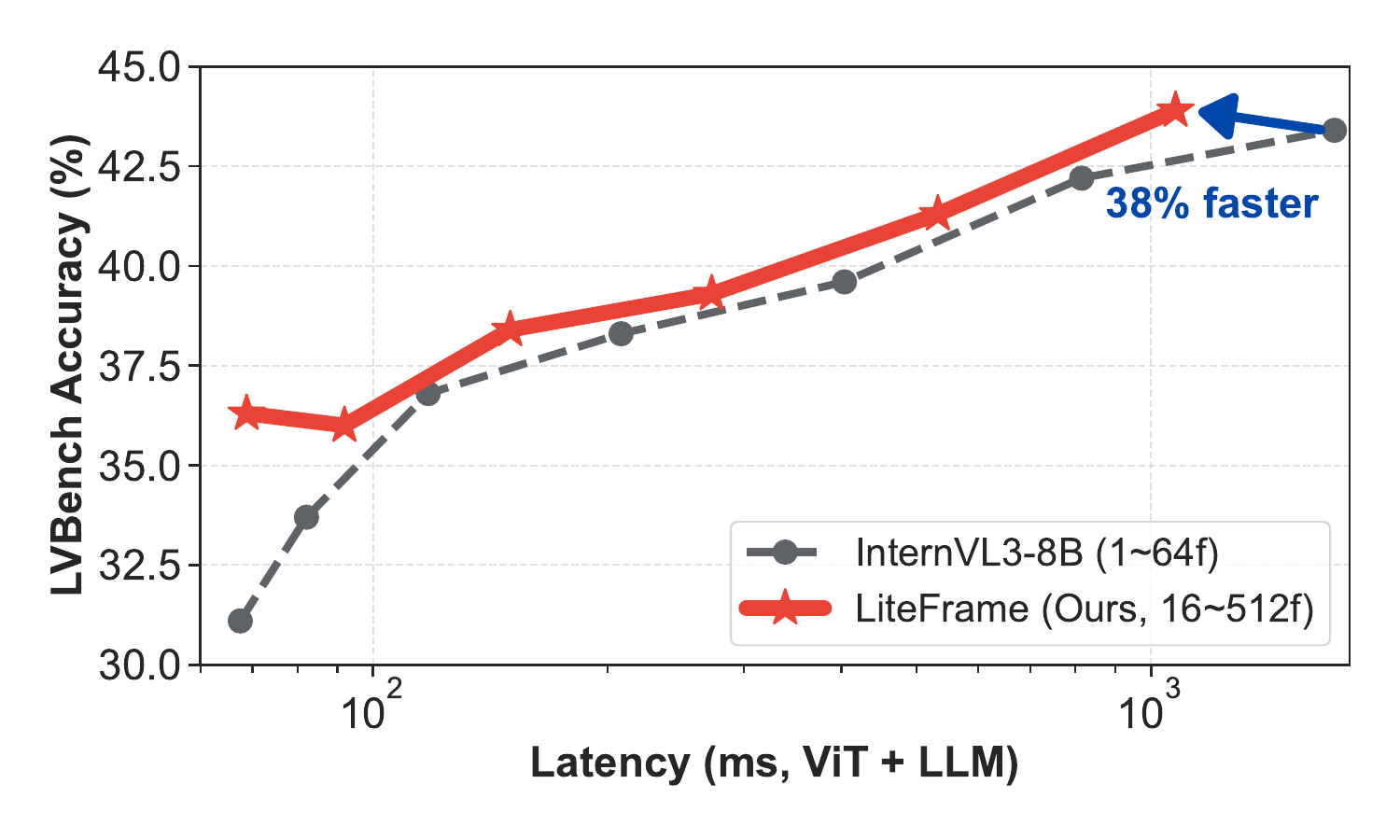} 
    \end{minipage}\hfill
    \begin{minipage}[b]{0.48\textwidth}
        \centering
        \includegraphics[width=\linewidth]{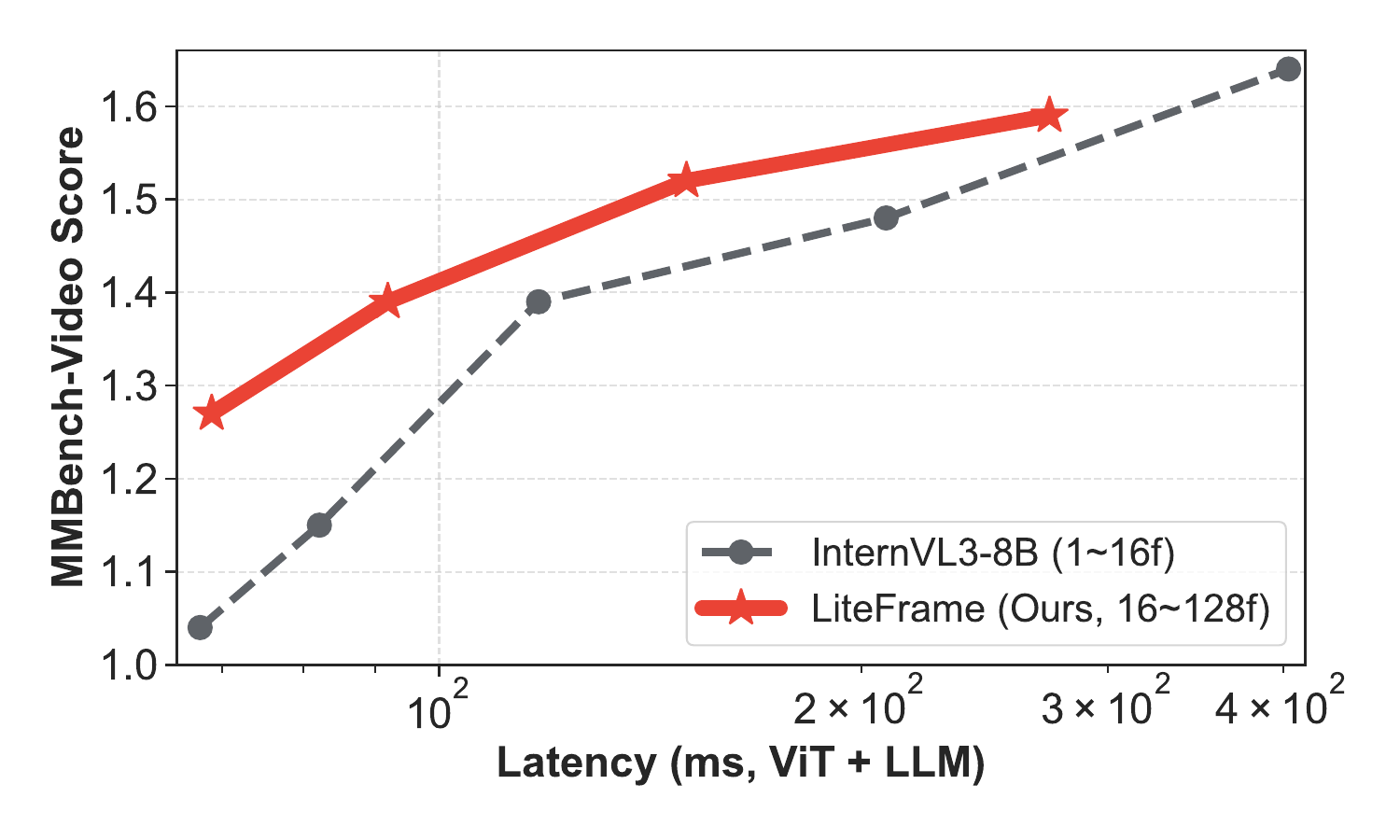} 
    \end{minipage}
    
    \vspace{2mm} 

    \caption{
        \textbf{Latency-accuracy trade-offs on long video benchmarks.}
        Additional evaluation results on LVBench~\textbf{(Left)} and MMBench-Video~\textbf{(Right)}.
        LiteFrame achieves the best score of 43.9 (vs.\ 43.5) on LVBench, while improves efficiency within the low-latency region on MMBench-Video.
    }
    \label{fig:lvbench_mmbenchvideo}
\end{figure}

\newpage
\section{Details on ablation studies}
\label{sec:app_ablation}

\subsection{Reconstructive training objective}
\label{subsec:ablation_training}

While Weighted Average Pooling~(WAP) serves as our primary and highly effective mechanism for token compression,
we explore whether a purely learned compression paradigm could surpass distilling WAP.
To this end, we introduce Reconstructive Token Distillation~(RTD), an exploratory variant that removes the constraints of a pre-defined latent space.
Instead, RTD employs an autoencoding objective where the student acts as the encoder and lightweight auxiliary transformer blocks serves as the decoder~($\text{Dec}$).
The exact objective is to reconstruct the teacher’s full dense feature map $T(x)$ from the student’s compressed latent representation $S(x)$:
$$\mathcal{L}_{\text{RTD}} = || T(x) - \text{Dec}(S(x)) ||^2_2$$

This objective encourages the student to learn a compression manifold that preserves the maximum amount of general visual information from the teacher, theoretically allowing the network to discover non-trivial spatio-temporal dependencies.

\begin{figure*}[!h]
    \centering
    \label{fig:ctd}
    \includegraphics[width=0.7\linewidth]{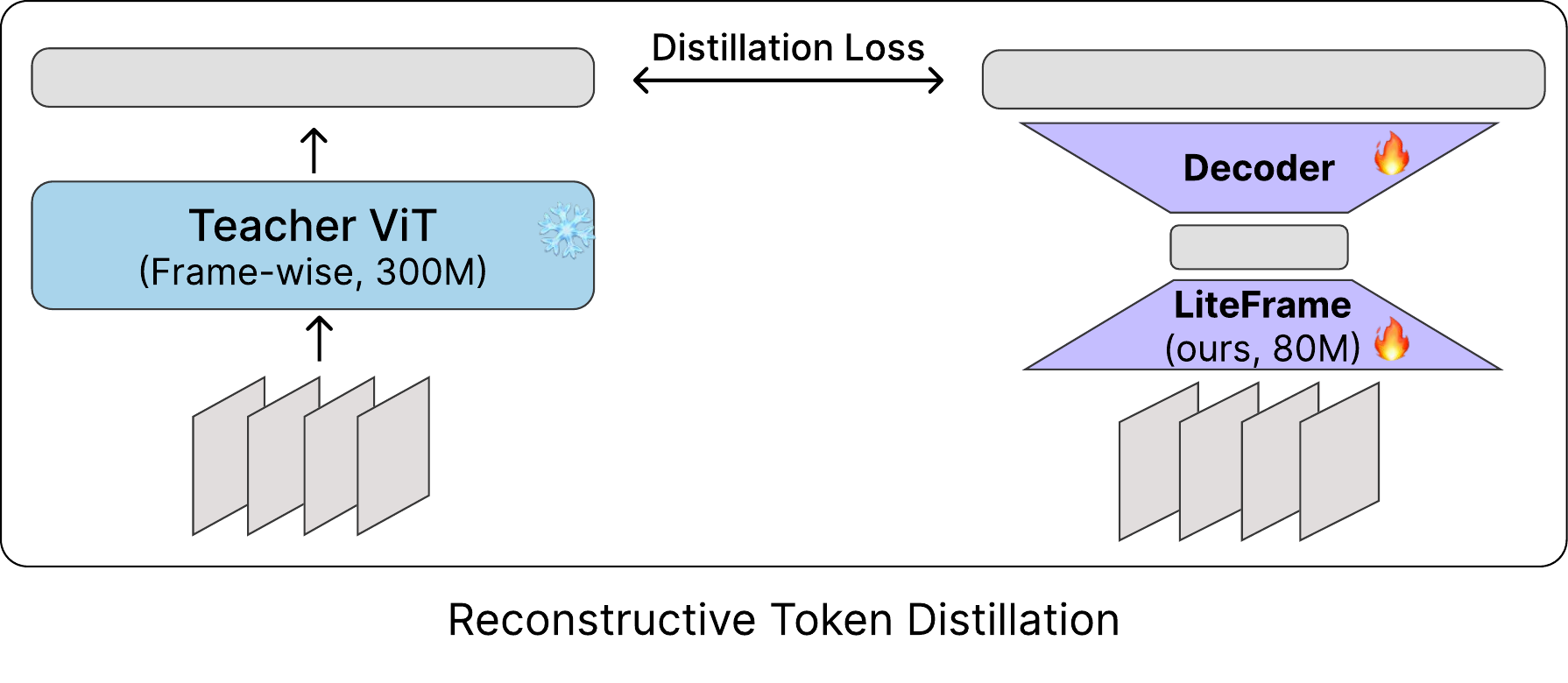}
    \caption{
    \textbf{Reconstructive Token Distillation~(RTD).}
    RTD enables the student to learn a compressed latent space via an auxiliary auto-encoding strategy, serving as an ablation to our primary WAP-based distillation.
}
\end{figure*}

However, as shown in \Cref{tab:ablation_variants}, empirical results strongly validate our primary design choice~(i.e.\ CTD).
While the learnable compression~(RTD) yields competitive results when paired with LMA, CTD consistently achieves superior performance.
Notably, CTD without LMA already surpasses RTD with LMA.
This suggests that explicitly aligning the student with the WAP primitive provides a much more robust, task-relevant semantic foundation than a generic reconstruction objective.

Furthermore, the combination of CTD and LMA delivers the highest accuracy across all frame budgets, reaching 65.3\% at 256 frames.
This underscores a critical synergy between CTD and LMA.
CTD effectively distills the high-saliency feature maps of the teacher into the student’s weights, while the subsequent lightweight fine-tuning phase~(LMA) is essential for properly aligning the pretrained LLM with the spatio-temporally compressed, information-dense representations produced by our student encoder.

\begin{table}[!t]
    \centering
    \caption{\textbf{Ablation analysis of compressive token distillation variants.} 
    We evaluate the average performance~(\%) of different distillation strategies. \textbf{CTD}: Compressed Token Distillation; \textbf{RTD}: Reconstructive Token Distillation; \textbf{LMA}: Language Model Adaptation. \textbf{CTD + LMA} achieves the best overall trade-off.}
    \label{tab:ablation_variants}
    \resizebox{0.5\linewidth}{!}{
        \begin{tabular}{l c c c}
            \toprule
            & \multicolumn{3}{c}{\textbf{Average Accuracy (\%)}} \\
            \cmidrule(lr){2-4}
            \textbf{Method} & \textbf{64 frames} & \textbf{128 frames} & \textbf{256 frames} \\
            \midrule
            \textbf{RTD} & 43.6 & 43.8 & 43.9 \\
            \textbf{RTD + LMA} & 59.4 & 61.5 & 63.1 \\
            \textbf{CTD}      & 60.0& 62.8& 64.1\\
            \textbf{CTD + LMA} & \textbf{61.0}& \textbf{63.4}& \textbf{65.3}\\
            \bottomrule
        \end{tabular}
    }
\end{table}

\subsection{Distillation without compression}
To isolate the performance gains attributable to our token-compressive student from those of simple model distillation, we compare LiteFrame against a standard baseline where the heavy teacher is distilled into a ViT-Base-12L.
For a fair comparison, we evaluate performance without subsequent LMA, keeping the LLM frozen.
As demonstrated in \Cref{tab:ablation_strategies}, our ``Distill~(No Comp.)'' baseline suffers from the exact same latency paradox as the teacher: despite a drastic reduction in encoder computational cost~(18.5 ms vs.\ 40.0 ms at 8 input frames), the absence of token compression forces the LLM to process an excessive volume of visual tokens~(256 per frame), which severely bottlenecks overall efficiency.
Consequently, when constrained to a fixed latency budget, the uncompressed student is forced to process significantly fewer frames, resulting in suboptimal performance.

In contrast, LiteFrame~(CTD) drastically reduces the visual token volume to just 16 tokens per frame.
This efficiency offloads the critical prefilling bottleneck from the LLM, unlocking the capability to process $8\times$ more frames within much lower latency (272.6ms vs.~393.3ms at 256 input frames for CTD).

\subsection{Spatio-temporal vs. Spatial-only compression}

\begin{figure}[!h]
    \centering
    \label{fig:spatial_only}
    \includegraphics[width=0.55\linewidth]{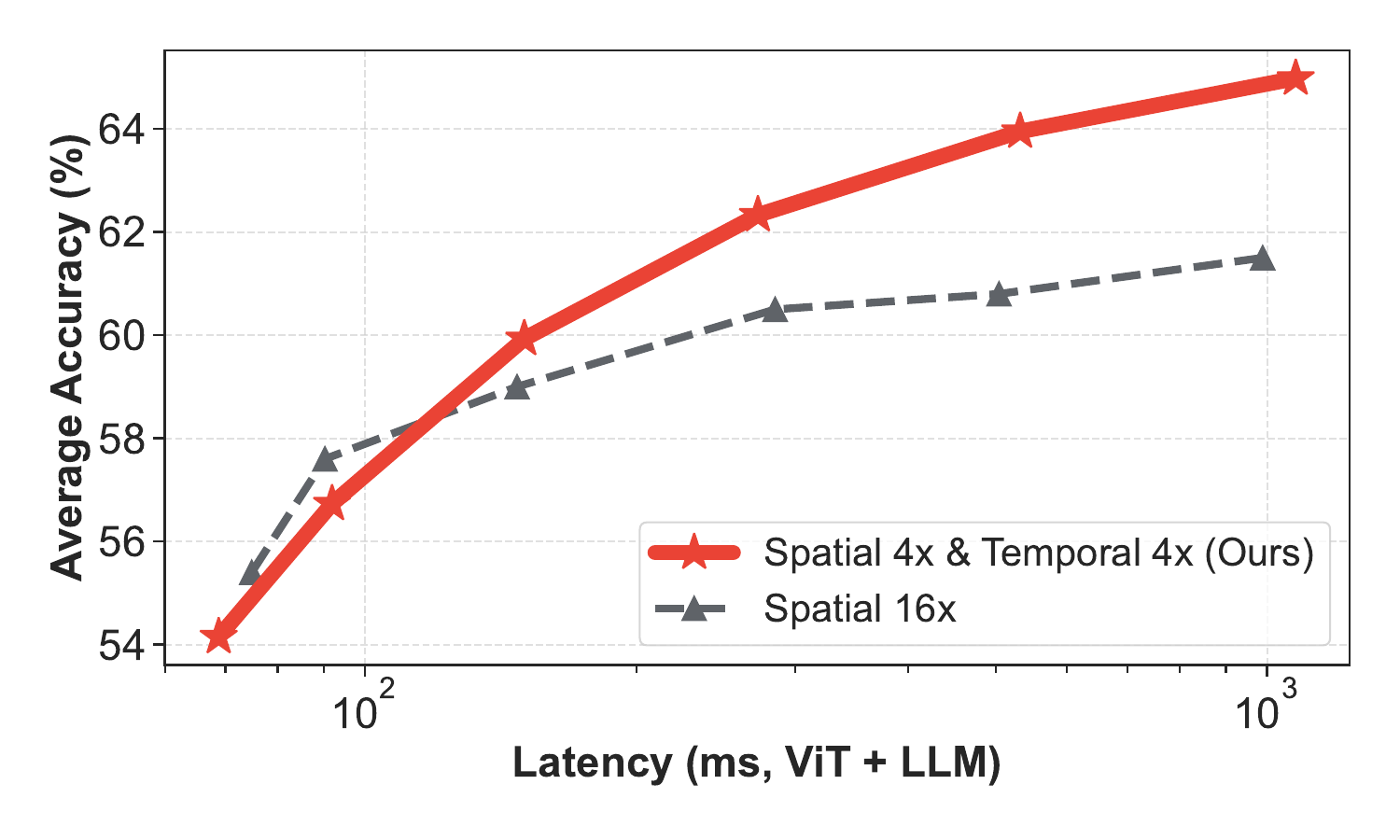}
    \caption{\textbf{Comparison between spatio-temporal compression and spatial-only compression.}
    We compare the accuracy of $16\times$ spatial-only compression and spatio-temporal compression~($4\times$ spatial and $4\times$ temporal).
    Spatio-temporal compression consistently outperforms spatial-only compression by effectively eliminating temporal redundancies between adjacent frames while preserving spatial fidelity.
}
\end{figure}

We further scrutinize the necessity of temporal compression by comparing our proposed spatio-temporal reduction~(Spatial $4\times$ and Temporal $4\times$) against a spatial $16\times$ baseline.
This baseline applies internal pooling solely along the spatial dimension without temporal layers~(i.e.\ ImageViT), and is trained to predict the teacher's compressed features using spatial $16\times$ WAP.

While the Spatial $16\times$ variant theoretically adheres to the same overall token budget, it significantly underperforms our spatio-temporal approach across most of the latency region.
Furthermore, the performance gap widens as the input frame count increases, as shown in \Cref{fig:spatial_only}.

Specifically, at the 128-frame budget, the spatial-only baseline achieves 60.5\% average accuracy compared to our 62.8\%, with a notable drop in fine-grained spatial understanding required for rigorous benchmarks like Video-MME~(57.0\% vs.\ 61.9\%).
We attribute this degradation to the excessive loss of spatial fidelity required to meet the compression target~(e.g., destructively pooling a frame into a $4\times 4$ grid).
By distributing the compression load across both spatial and temporal dimensions, our approach preserves critical spatial details while effectively aggregating redundant temporal dynamics, resulting in a far more balanced and information-rich representation for downstream LLM reasoning.

\begin{table*}[!t]
\centering
\caption{\textbf{Ablation analysis of compression strategies.} Comparison of our proposed spatio-temporal compression (Ours) against basic distillation without compression (ViT-Base-12L) and spatial-only compression. The best results are marked in \textbf{bold}. \textbf{Ours (CTD)} consistently outperforms the others.} 
\vspace{4mm}
\label{tab:ablation_strategies}
\resizebox{\textwidth}{!}{
\begin{tabular}{lccccccccccc}
\toprule
\multirow{2}{*}{Method} & Vision & Tokens/ & \multirow{2}{*}{Frames} & \multicolumn{3}{c}{Latency (ms)} & \multicolumn{5}{c}{Accuracy (\%)} \\
\cmidrule(lr){5-7} \cmidrule(lr){8-12}
& Params & Frame & & Vision & LLM & Total & V-MME$^{\dagger}$ & V-MME$^{\ddagger}$ & MLVU & LongVideo & Avg \\
\midrule
\rowcolor{gray!10} InternVL3-8B & 304M & 256 & 8 & 40.0 & 167.3 & 208.4 & \textbf{59.0} & 60.0 & 62.2 & 54.8 & 59.0 \\
Distill (No Comp.) & 86M & 256 & 8 & 18.5 & 166.5 & 185.2 & 55.9 & 58.1 & 58.6 & 53.8 & 56.6 \\
Spatial 16x & 86M & 16 & \textbf{64} \scriptsize{(8$\times$)} & 49.8 & 96.2 & \textbf{147.4} & 56.2 & 62.0 & 63.2 & 54.6 & 59.0 \\
\textbf{Ours (CTD)} & \textbf{87M} & 16 & \textbf{64} \scriptsize{(8$\times$)} & 54.8 & 94.9 & 150.1 & 58.7 & \textbf{62.1} & \textbf{65.1} & 53.9 & \textbf{60.0} \\
\midrule
\rowcolor{gray!10} InternVL3-8B & 304M & 256 & 16 & 74.0 & 329.3 & 403.6 & \textbf{61.9} & 64.0 & \textbf{66.4} & 56.5 & 62.2 \\
Distill (No Comp.) & 86M & 256 & 16 & 34.3 & 328.6 & 362.9 & 59.4 & 61.4 & 64.5 & 55.7 & 60.3 \\
Spatial 16x & 86M & 16 & \textbf{128} \scriptsize{(8$\times$)} & 93.4 & 166.5 & \textbf{264.8} & 57.0 & 64.1 & 64.6 & 56.4 & 60.5 \\
\textbf{Ours (CTD)} & \textbf{87M} & 16 & \textbf{128} \scriptsize{(8$\times$)} & 105.3 & 166.6 & 272.6 & \textbf{61.9} & \textbf{65.6} & 65.8 & \textbf{57.7} & \textbf{62.8} \\
\midrule
\rowcolor{gray!10} InternVL3-8B & 304M & 256 & 32 & 144.5 & 669.8 & 814.5 & \textbf{65.6} & \textbf{67.4} & \textbf{69.6} & \textbf{58.7} & \textbf{65.3} \\
Distill (No Comp.) & 86M & 256 & 32 & 63.9 & 670.0 & 733.8 & 60.6 & 64.2 & 67.5 & 58.2 & 62.6 \\
Spatial 16x & 86M & 16 & \textbf{256} \scriptsize{(8$\times$)} & 185.2 & 319.1 & \textbf{504.4} & 57.2 & 64.7 & 65.4 & 55.8 & 60.8 \\
\textbf{Ours (CTD)} & \textbf{87M} & 16 & \textbf{256} \scriptsize{(8$\times$)} & 204.0 & 327.4 & 532.3 & 62.7 & 67.3 & 68.8 & 57.7 & 64.1 \\
\bottomrule
\end{tabular}
}
\end{table*}

\newpage
\section{Detailed comparisons with AutoGaze}
\label{sec:app_autogaze}

\begin{figure}[!t]
    \centering
    \begin{minipage}[c]{0.35\linewidth}
        \centering
        \includegraphics[width=\linewidth]{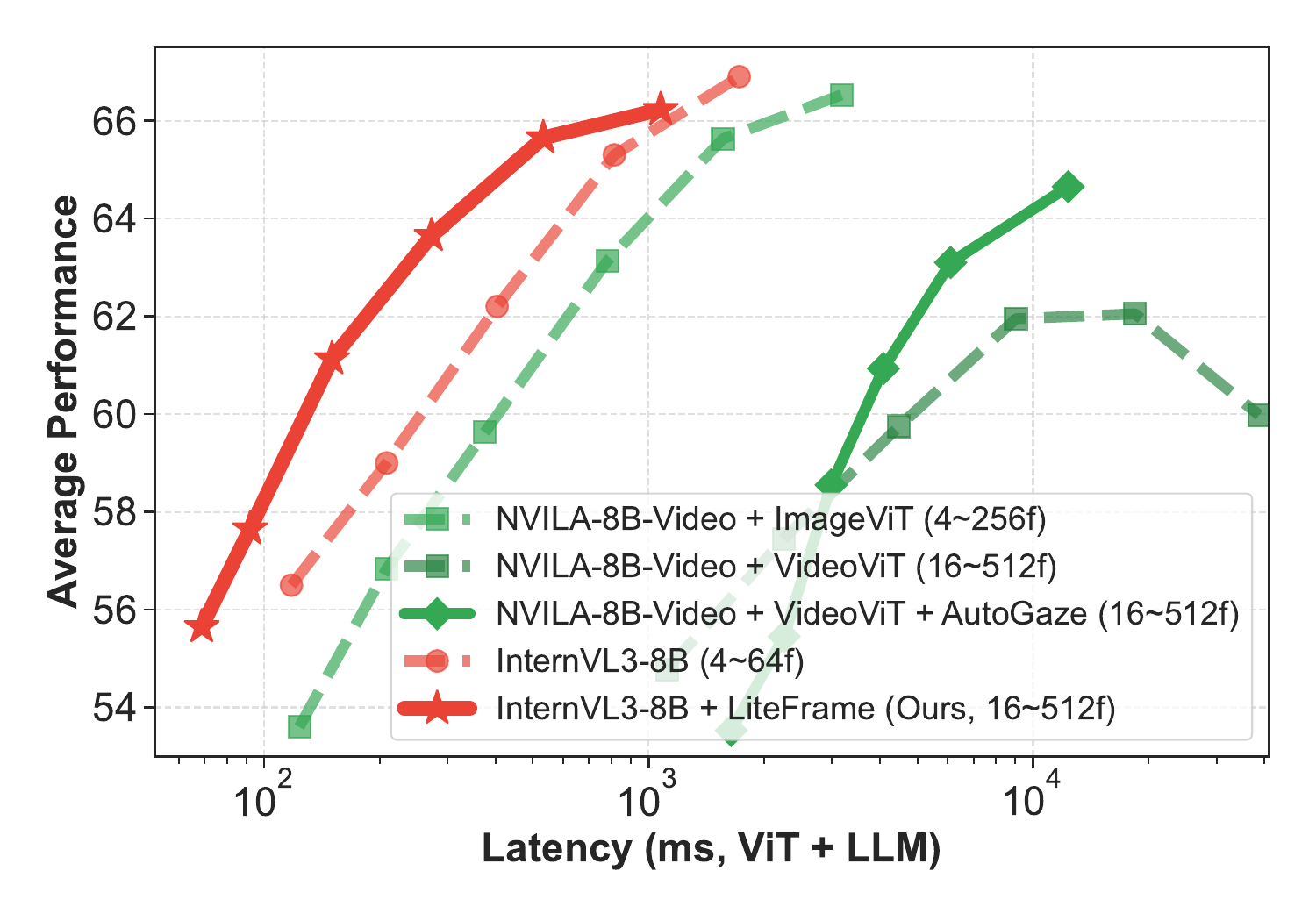}
    \end{minipage}\hfill
    \begin{minipage}[c]{0.63\linewidth}
        \vspace{-5mm}
        \centering
        \resizebox{\linewidth}{!}{
        \begin{tabular}{lcccccc}
        \toprule
        \multirow{2}{*}{Method} & \multirow{2}{*}{Frames} & \multicolumn{4}{c}{Latency (ms)} & \multirow{2}{*}{Avg. (\%)} \\
        \cmidrule(lr){3-6}
         & & AutoGaze & ViT & LLM & Total & \\
        \midrule
        \rowcolor{gray!10} NVILA-8B-Video~(ImageViT) & 32 & - & 451.0 & 329.9 & 780.8 & 63.1 \\
        \rowcolor{gray!10} NVILA-8B-Video~(VideoViT) & 32 & - & 1916.0 & 330.8 & 2246.8 & 57.5 \\
        \begin{tabular}{@{}c@{}}+AutoGaze~(VideoViT) \\ \scriptsize{}{(compared with ImageViT)}\end{tabular} & \begin{tabular}{@{}c@{}}256 \\ \scriptsize{}{(8$\times$)}\end{tabular} & \begin{tabular}{@{}c@{}}2961.4 \\ \neginc{+2961.4ms}\end{tabular} & \begin{tabular}{@{}c@{}}2605.6 \\ \neginc{+477.8\%}\end{tabular} & \begin{tabular}{@{}c@{}}539.5 \\ \neginc{+63.5\%}\end{tabular} & \begin{tabular}{@{}c@{}}6106.5 \\ \neginc{+682.1\%}\end{tabular} & \begin{tabular}{@{}c@{}}63.1 \\ \rem{0.00}\end{tabular} \\
        \midrule
        \rowcolor{gray!10}InternVL3-8B & 32 & - & 144.5 & 669.8 & 814.5 & 65.3 \\
        +LiteFrame~\scriptsize{(Ours)} & \begin{tabular}{@{}c@{}}256 \\ \scriptsize{}{(8$\times$)}\end{tabular} & - & \begin{tabular}{@{}c@{}}{204.0} \\ \neginc{+41.2\%}\end{tabular} & \begin{tabular}{@{}c@{}}{327.4} \\ \posdec{-51.1\%}\end{tabular} & \begin{tabular}{@{}c@{}}\textbf{532.3} \\ \posdec{-34.6\%}\end{tabular} & \begin{tabular}{@{}c@{}}\textbf{65.7} \\ \inc{+0.4}\end{tabular} \\
        \bottomrule
        \end{tabular}
        }
    \end{minipage}

    \vspace{-4mm}

    \caption{
        \textbf{Detailed comparison with AutoGaze.} 
        \textbf{(Left)} LiteFrame~(red) improves the efficiency frontier with higher accuracy and lower latency, whereas AutoGaze~(green) introduces  auxiliary overhead compared with standard VLMs with ImageViT.
        \textbf{(Right)} When scaling from 32 to 256 frames ($8\times$), LiteFrame balances ViT and LLM scaling, while AutoGaze's pre-reduction module becomes a new bottleneck. Note: Relative changes for +AutoGaze are computed against the standard ImageViT baseline, highlighting the architectural overhead.
    }
    \label{fig:app_movinet_autogaze_combined}
\end{figure}

\subsection{Detailed comparison}

We compare our method against AutoGaze~\citep{shi2026autogaze}, a recent approach that similarly addresses the dual computational bottlenecks of the ViT and LLM.
We benchmark the latency-performance trade-offs when integrating LiteFrame and AutoGaze into their respective baselines.
It is important to note that AutoGaze structurally requires a VideoViT backbone~(full spatio-temporal attention across 16-frame clips), whereas standard VLMs utilizes a much lighter ImageViT.
For a transparent comparison, we report two NVILA-8B-Video baseline using ImageViT and VideoViT.
As illustrated in \Cref{fig:app_movinet_autogaze_combined}~(left), while AutoGaze successfully accelerates its own heavy VideoViT baseline, it remains substantially slower than a standard baseline equipped with an ImageViT.
The detailed breakdown in \Cref{fig:app_movinet_autogaze_combined}~(right) reveals that the AutoGaze pre-reduction module introduces a severe latency bottleneck, accounting for nearly half of total inference time~(3.0s out of 6.1s).
In contrast, LiteFrame is designed directly for the standard VLM paradigm, without incurring auxiliary overheads.
This allows LiteFrame to strictly advance the Pareto frontier, lowering the total latency without compromising performance.

\subsection{Evaluation setups for AutoGaze}
For evaluating AutoGaze, we employ the optimal set of hyperparameters provided by the authors: $\text{task\_loss\_requirement\_tile}=0.6$, $\text{gazing\_ratio} = [1] + [0.3] * 15$ and $\text{target\_scales} = [64,128,224,448]$.
In addition, we fix the input resolution to 448px to directly match the input of our method, and we set $\text{num\_video\_frames\_thumbnail}=\text{num\_video\_frames} // 16$ to avoid using excessive thumbnail frames.

To benchmark the latency, we explicitly exclude video preprocessing overhead of AutoGaze~(reading the video, constructing the image pyramid, and resizing frames).
This ensures a fair comparison that focuses only on the neural-network execution time.
Since AutoGaze's latency varies across videos, we measure the median latency on Video-MME.

Specifically, NVILA-8B-Video~(both ImageViT and VideoViT) refer to the NVILA-HD-8B-Video checkpoints evaluated without incorporating AutoGaze module (i.e.~setting $\text{task\_loss\_requirement\_tile}=1.0$).
The ImageViT variant encodes videos in a frame-wise manner, whereas VideoViT variant concatenates all visual tokens from a 16-frame clip and proess them through a full spatio-temporal transformer.

\section{Limitations}
\label{sec:app_limitations}
While LiteFrame establishes a new efficiency-accuracy Pareto frontier for video understanding, we acknowledge a few limitations that present promising directions for future work.
First, our Language Model Adaptation~(LMA) was trained using a subset of existing video data.
Incorporating higher-quality, extreme long-form video datasets could potentially maximize the benefits of our extended temporal context window, further elevating performance without requiring any architectural changes to our core contribution.
Second, because our primary focus is mitigating the temporal scaling paradox and frame-count bottlenecks inherent to Video LLMs, we evaluated LiteFrame exclusively on video-centric benchmarks.
Although the model exhibits promising zero-shot spatial scaling capabilities, its performance on purely static, traditional image benchmarks remains unexplored.
Finally, while we successfully distilled the vision encoder from a heavy 304M parameters teacher into a lightweight 87M parameters student, efforts to scale down to even smaller student models were constrained by training instabilities, such as loss explosions.
Advancing the Compressed Token Distillation~(CTD) framework to stabilize the training of ultra-lightweight students remains a highly promising next step.


\end{document}